\documentclass[preprint,11pt]{elsarticle}
\usepackage[latin1]{inputenc}
\usepackage[english]{babel}

\usepackage{amsmath,amssymb}
\usepackage{bbm}
\usepackage{graphbox}
\usepackage{graphicx}
\usepackage{setspace}
\usepackage[table]{xcolor}
\usepackage{microtype}
\usepackage{multirow}
\usepackage{enumitem}
\usepackage{algorithm}
\usepackage{algpseudocode}
\usepackage{subfigure}
\usepackage{booktabs}
\usepackage{tikz}
\usepackage{xcolor}
\usepackage{hyperref}

\newcommand{\Mu}{M}
\renewcommand{\vec}[1]{\boldsymbol{#1}}

\newcommand{\bigO}[1]{\mathcal{O}({#1})}
\newcommand{\tableref}[1]{Table~\ref{#1}}
\newcommand{\figureref}[1]{Figure~\ref{#1}}
\newcommand{\equationref}[1]{Equation~(\ref{#1})}
\newcommand{\algorithmref}[1]{Algorithm~\ref{#1}}
\newcommand{\sectionref}[1]{Section~\ref{#1}}
\renewcommand{\algorithmiccomment}[1]{\bgroup\hfill\footnotesize~#1\egroup}

\DeclareMathOperator{\Tr}{Tr}

\newcommand{\myblue}[1]{#1}

\allowdisplaybreaks
\title{Regularization and \myblue{Optimization} in Model-Based Clustering}

\author{Raphael Araujo Sampaio, Joaquim Dias Garcia, Marcus Poggi, Thibaut Vidal}

\begin{document}

\linespread{1.5}\selectfont

\begin{center}

\vspace*{-0.5cm}

\begin{huge}
Regularization and \myblue{Optimization} in\\
Model-Based Clustering
\end{huge}

\vspace*{0.6cm}

\textbf{Raphael Araujo Sampaio$^{a,b*}$, Joaquim Dias Garcia$^{a,c}$, Marcus Poggi$^{b}$, Thibaut Vidal$^{b,d}$} \\
$^a$ PSR, Rio de Janeiro, Brazil \\
$^b$ Departamento de Inform\'{a}tica, Pontif\'{i}cia Universidade Cat\'{o}lica do Rio de Janeiro (PUC-Rio), Brazil \\
$^c$ LAMPS, Departamento de Engenharia El\'{e}trica, PUC-Rio, Brazil \\
$^d$ CIRRELT \& SCALE-AI Chair in Data-Driven Supply Chains, Department of Mathematical and Industrial Engineering, Polytechnique Montr\'{e}al, Canada \\
rsampaio@psr-inc.com, joaquim@psr-inc.com, poggi@inf.puc-rio.br, thibaut.vidal@polymtl.ca\\
\vspace*{0.15cm}

\end{center}
\noindent
\textbf{Abstract.}
Due to their conceptual simplicity, k-means algorithm variants have been extensively used for unsupervised cluster analysis. However, one main shortcoming of these algorithms is that they essentially fit a mixture of identical spherical Gaussians to data that vastly deviates from such a distribution. In comparison, general Gaussian Mixture Models (GMMs) can fit richer structures but require estimating a quadratic number of parameters per cluster to represent the covariance matrices. This poses two main issues: (i) the underlying optimization problems are challenging due to their larger number of local minima, and (ii) their solutions can overfit the data. In this work, we design search strategies that circumvent both issues. We develop more \myblue{effective optimization} algorithms for general GMMs, and we combine these algorithms with regularization strategies that avoid overfitting. Through extensive computational analyses, we observe that optimization or regularization in isolation does not substantially improve cluster recovery. However, combining these techniques permits a completely new level of performance previously unachieved by k-means algorithm variants, unraveling vastly different cluster structures. These results shed new light on the current status quo between GMM and k-means methods and suggest the more frequent use of general GMMs for data exploration. To facilitate such applications, we provide open-source code as well as Julia packages (\textsc{UnsupervisedClustering.jl} and \textsc{RegularizedCovarianceMatrices.jl}) implementing the proposed techniques.
\vspace*{0.2cm}

\vspace*{-0.5cm}

\noindent
$^*$ Corresponding author

\linespread{1.8}\selectfont

\section{Introduction}
\label{section:introduction}

Unsupervised clustering is the process of splitting unlabeled data into homogeneous groups with similar samples. This task is essential for data preparation and exploration. Moreover, it is linked with numerous applications in social media analysis, image processing, text analysis, and bioinformatics, among many others \citep{Jain2010}. As a consequence of the diversity of applications, data sources, similarity measures, goals, and constraints, clustering encompasses a vast universe of methods.

Model-based clustering approaches hold a central place due to their well-grounded statistical principles and their controllable parsimony \citep{Bouveyron2014,Mcnicholas2016}. These approaches assume that each cluster corresponds to a statistical process represented by a probability density function (e.g., mixtures of multivariate Gaussians). Maximum likelihood estimators of the statistical processes are used to obtain the cluster labels, which are unobserved latent parameters.

Restricting the Gaussian mixtures to be spherical (identity matrix as covariance) leads to a maximum likelihood problem almost equivalent to the minimum sum-of-squares clustering problem \citep{Bishop2006}. The classical k-means algorithm \citep{hartigan1979k, lloyd1982least} is a well-known local optimizer for this problem. It optimizes the $\bigO{Kd}$ model parameters representing the mean (or centroid) of $K$ Gaussian distributions (i.e., the center of each cluster in a feature space of dimension~$d$). The covariance matrix of these Gaussians is the identity matrix in the case of k-means. Due to their speed and simplicity, k-means and their variations have been widely used. 

Assuming only spherical Gaussian mixtures is a rough approximation of real datasets. More general models with arbitrary covariance matrices have been studied to better fit the data, leading to \emph{elliptical} variants of the k-means algorithms \citep{cerioli2005k,xiang2008learning,morales2014fast} and the well-known expectation-maximization (EM) clustering algorithms \citep{dempster1977maximum,mclachlan2007algorithm} typically used to search for the maximum likelihood of the Gaussian mixture.

Although more general, the joint estimation of covariances and means within a Gaussian mixture model poses at least two fundamental issues. Firstly, the covariance matrices can become ill-conditioned during likelihood maximization without proper regularization. The ill-conditioned matrices create numerical instabilities, prevent matrix inversion, and increase the risk of overfitting the data instead of revealing useful structure data. Secondly, due to its larger number of parameters, likelihood maximization is much harder to solve in a general form, and the presence of numerous local maxima cripples solution methods. The difficulty of this optimization problem is the main reason why general EM and elliptical k-means variants have been previously qualified as a ``failed opportunity'' in the clustering domain \citep{cerioli2005k}.

In this study, we revisit some of these methods, considering different regularization techniques and \myblue{optimization} methods, to better understand to what extent and with which components general EM can be effectively used. Indeed, significant progress has been made in covariance matrix estimation, and many regularized estimators have been proposed. We attempt to use such sophisticated regularized estimators within EM-based algorithms. However, despite these regularization techniques, we observe that general EM models' performance remains underwhelming in feature spaces of dimensions 20 or more. This lack of performance apparently relates to the second issue, i.e., the presence of numerous local maxima. To circumvent this issue, we extend previous works focused on generating promising starting points for EM algorithms \citep{Biernacki2003} and propose more sophisticated  \myblue{optimization metaheuristics} relying on recombination operations inspired by the study of \cite{gribel2019hg}. As visible in our experiments, this second improvement counterbalances the weaknesses of the general EM, leading to high-quality solutions in a much larger regime for datasets with fewer samples relative to the feature space dimension.

To summarize, the contributions of this paper are the following.
\begin{itemize}
\item We test the general EM model with different regularization strategies, identifying the most successful approaches in terms of solution quality and computational effort.
\item We propose advanced \myblue{optimization} algorithms that permit an escape from local maxima and effectively combine them with the regularization strategies. The proposed methods can be viewed as a multi-start EM in which the initial parameters of the clusters (covariance and mean) are carefully inherited and recombined from previously found solutions.
\item We demonstrate, on controlled synthetic data, that the \emph{joint} use of sophisticated regularization and \myblue{optimization} techniques allows for the recovery of the original clusters much more systematically, outperforming classical spherical EM or k-means on a wider range of datasets.
\item We compare the proposed methods in real data sets and verify that the regularized methods with \myblue{optimization} achieve better cluster recovery.
\end{itemize}

This paper is structured as follows. \sectionref{section:review} discusses fundamental notions and reviews the related literature on model-based clustering, focusing principally on regularization and \myblue{optimization}. \sectionref{section:methodology} presents the proposed methodology, including a \myblue{optimization} technique and different regularization methods. \sectionref{section:experiments} presents our computational experiments on synthetic and real datasets. Finally, \sectionref{section:conclusion} concludes and points to directions for future research.

\section{Fundamental Notions and Related Studies}
\label{section:review}

Let $X$ be a dataset containing $n$ data points (i.e., samples), $X = \{\vec{x}_1, ..., \vec{x}_n\}$, in which each point $\vec{x}_i = \{x_{i}^{(1)}, ..., x_{i}^{(d)}\}$ is represented as a vector of $d$ features. We consider an unsupervised learning context in which the samples do not have known labels (i.e., groups), and the only available information is the number of these $k$ labels.

We aim to find an underlying structure in $X$ in the form of clusters. More precisely, we should distribute these samples into $k$ labels so that the samples that are most similar to each other belong to the same cluster, and those with low similarity belong to different clusters.
 
Clustering algorithms can be broadly divided into two leading families: \emph{soft} assignment methods output points-to-cluster assignments in the form of a probability distribution, whereas \emph{hard} assignment methods associate each data point to a single cluster. Moreover, since clustering is a very general task, there exists, within each sub-category, various approaches, which can lead to the discovery of different solutions and underlying structures \citep{estivill2002so}.\\

\noindent
\textbf{Expectation Maximization.}
Among soft assignment clustering methods, Expectation-Maximization (EM) algorithms based on Gaussian Mixture Models (GMM) hold a prominent place due to their generality and simplicity. Proposed by \citet{dempster1977maximum}, EM is a general solution approach for maximum likelihood estimation (MLE) problems with missing data \citep[also see][]{bilmes1998gentle,blum2016foundations,mclachlan2007algorithm}. EM jointly estimates model parameters and missing data (the sample label in the case of clustering). In its most general form, EM searches for the model parameters $\theta$ that maximize the expected value of a log-likelihood function, $\log \mathcal{L}(\theta | X, Y)$, which depends on available information, $X$, and hidden information, $Y$, about the samples.
$Y$ has a posterior probability distribution as a function of $X$ and $\theta$, i.e., $Y | X, \theta$. Therefore, we take the expectation concerning $Y$ conditioned on both $X$ and $\theta$:
\begin{align}
\max_\theta \quad \mathop{\mathbb{E}}_{Y | X, \theta} [\log \mathcal{L}(\theta | X, Y)].
\label{em:maxexpected}
\end{align}

As their name indicates, EM algorithms include two steps: \emph{expectation} and \emph{maximization}. In the expectation step, parameters $\theta$ are fixed to calculate a posterior of $Y$ given $X$ and $\theta$. Consequently, we can obtain the expected log-likelihood function $\theta \rightarrow Q(\theta)$. In any iteration $t$, $Q^{(t)}(\theta)$ is obtained by fixing a previous value of the parameters $\theta^{(t-1)}$ in the expectation conditioning, and it is expressed as:
\begin{align}
Q^{(t)}(\theta) = \mathop{\mathbb{E}}_{Y | X, \theta^{(t-1)}} \left[ \log \mathcal{L}(\theta | X, Y) \right].
\label{em:eq1}
\end{align}
Based on this, the maximization step optimizes the current approximation $Q^{(t)}(\theta)$ with respect to $\theta$ to obtain an improved estimate, as shown in \equationref{em:eq2}:
\begin{align}
\theta^{(t)} = \underset{\theta}{\text{argmax}} \ Q^{(t)}(\theta).
\label{em:eq2}
\end{align}
 
The method, therefore, iteratively attempts to estimate the model parameters $\theta$ and the labels $Y$. In each step, the likelihood function is improved. These steps are repeated until a local optimum is attained.\\

\noindent
\textbf{Gaussian Mixture Models.}
One frequently used generative model for clustering is to assume that data is distributed as a mixture of Gaussian distributions, i.e., a Gaussian Mixture Model (GMM --  \citealt{banfield1993model,moore1999very,zivkovic2004improved}). GMM is a generative process involving two steps for each sample: i) deciding from which cluster the sample will belong; ii) generating the sample from the corresponding Gaussian distribution of the mixture. Each Gaussian is characterized by a mean vector, $\vec{\mu}$, and a covariance matrix, $\Sigma$. The vector $\vec{\pi}$ defines the weights of the Gaussians. The joint probability distribution of all samples is given by:
\begin{align}
\sum_{j=1}^{k} \pi_j \ \mathcal{N}(X \ | \ \vec{\mu}_j, \Sigma_j)
\label{mixture-of-gaussians}
\end{align}

In this model, $\vec{\mu}_j$ and $\Sigma_j$ can be seen as the parameters governing the shape of the $j^{th}$ cluster. The eigenvectors of $\Sigma_j$ describe the ellipsis' main axis orientation, while the eigenvalues describe their length. Finally, $\pi_j$ is the \textit{a priori} probability that a sample belongs to a given cluster~$j$.\\

\noindent
\textbf{Covariance estimation within EM-GMM.}
Estimating the covariance matrices in the EM-GMM gives greater flexibility in the families of data that can be accurately represented, but it also poses significant methodological challenges. In particular, the number of parameters of a covariance matrix grows quadratically with the number of features. Consequently, the number of samples involved in the estimate must be large enough to avoid over-fitting with ill-conditioned or singular matrices \citep{hawkins2004problem}. This also poses significant numerical issues in EM-type algorithms since full-rank matrices are typically needed in their intermediary steps.
To circumvent this issue, several studies have proposed regularization (or shrinkage) techniques \citep{ledoit2004honey}. Regularization of covariance matrices can improve the estimator by reducing the condition number, which is the ratio between the largest and the smallest eigenvalue.

Most covariance regularization methods return a convex combination of the empirical matrix $\Sigma$ with a scaled identity matrix, as shown in \equationref{eq:covariance-shrunk}.

\begin{align}
\Sigma_{\text{reg}} = (1-\delta) \Sigma + \delta\frac{\Tr\Sigma}{d}I
\label{eq:covariance-shrunk}
\end{align}

The constant multiplying of the identity matrix permits keeping the regularized matrix's magnitude similar to the magnitude of the original matrix. This constant is the average of the diagonal entries of $\Sigma$, equivalently written as $\frac{\Tr\Sigma}{d}$.
 
The empirical covariance matrix can be seen as a random variable having its own moments (e.g., variance and expectation).
If some changes in samples lead to large changes in the estimated matrix, then the estimator's variance is significant.
The original empirical estimator $\Sigma$ has a small bias but a large variance. On the other hand, $\frac{\Tr\Sigma}{d}I$ usually has a much smaller variance and a large bias. The purpose of regularization is to reduce the variance of estimators (at the cost of a higher bias). Therefore, the parameter $\delta \in [0,1]$ controls the trade-off between bias and variance. 

The regularization techniques considered in this work are all variants of the above-mentioned convex combination method, but they differ in the $\delta$ constant choice. This calculation also has immediate implications on the eigenvalues: the eigenvalues of $\Sigma_{\text{reg}}$ are eigenvalues of the empirical matrix $\Sigma$ to which a positive constant has been added. Therefore, indirectly, the regularization imposes a lower bound on the eigenvalues of the covariance matrix.

The ill-conditioned matrices estimating problem has inspired several studies. \citet{hastie1996discriminant} developed one of the first classifications works using mixtures of Gaussians and regularization techniques. Different approaches were used to calibrate the $\delta$ parameter and achieve a good covariance estimation, such as (i) maximizing the Leave-One-Out Likelihood (LOOL) criterion for each cluster \cite{dundar2002model}; (ii) estimating the parameter from the data using the Minimum Message Length (MML) principle \citep{law2004simultaneous}; and (iii) using a modified Bayesian Information Criterion (BIC) as a model selection criterion to determine $\delta$ \citep{pan2007penalized}.

The Ledoit-Wolf (LW) approach \citep{ledoit2004well} is a well-known regularization methodology that minimizes the mean squared error between the estimated and the true covariance matrix in the case of arbitrary distribution of the samples. The LW leads to well-conditioned matrices, and it is based on a simple \myblue{analytical} formula that can be easily and efficiently implemented. \citet{halbe2013regularized} used this technique on EM-GMM and noticed significant improvements in the performance of multivariate probability density estimation. Next, \cite{chen2010shrinkage} proposed another well-known method, called Oracle Approximating Shrinkage (OAS), which reduces the LW mean squared error for the special case of Gaussian processes.

\myblue{The research in} \cite{warton2008penalized} proposes and analyzes the selection of $\delta$ by cross-validation, leading to better results at the expense of a larger computational effort. \cite{won2013condition} describes a convex optimization problem minimizing a Gaussian log-likelihood function with an explicit constraint on the condition number of the matrix. Although the authors present an efficient algorithm, it is still bounded by a spectral decomposition. Finally, \cite{ledoit2019power} describes a non-linear regularization technique in which it is possible to select a specific regularization term for each eigenvalue leading to possibly better results than standard LW. However, like the previous method, it also requires a spectral decomposition.\\

\textbf{Finding better local optima in EM-GMM.} 
Finding good solutions to the likelihood maximization problem in general GMMs is challenging due to the large number of parameters that need to be estimated. Effectively, the most popular clustering algorithms are local search methods, such as k-means and EM-GMM, which can easily get trapped in local optima and depend on the initial conditions of the search. A common strategy to improve such methods consists of repeating the approach from different initial points and keeping the best solution. Although this is the first step toward methodological improvements, more efficient ways exist to improve the search using classical meta-heuristics, which are methods precisely designed to ``orchestrate the local search and higher-level strategies to escape from local optimum and reach better solutions'' \citep{glover2006handbook}.
 
A simple meta-heuristic approach, called Iterated Local Search (ILS) \citep{lourencco2003iterated}, consists of iteratively applying a local optimization algorithm (e.g., k-means or EM) to reach a local minimum, followed by a perturbation operator to generate a new starting point and pursue the search. This strategy effectively produces a diversity of starting points and profits from the information from previous iterations. As such, it outperforms random restarts in a variety of applications. One application of the ILS concept to clustering problems has been proposed by \citet{franti2018efficiency} under the name of the Random Swap (RS) approach. The design of this algorithm follows the classical ILS framework. Each time the local search converges towards a solution, the algorithm applies a \textsc{Swap} which consists of moving a random cluster's position to the location of an arbitrary point in the dataset. \citet{zhao2012random} combined the Random Swap and EM (RSEM) to improve EM-GMM clustering. In each RSEM iteration, a random Gaussian component is removed and relocated to a random data point location. The covariance matrix and weights are kept the same as the previous iteration in the Random Swap operation to preserve the cluster's parameters' relative magnitude.

Another way to search for a \myblue{superior optimized solution} is with genetic algorithms (GA). GAs rely on the basic principles of natural evolution (selection, mutation, crossover) to improve a population of solutions. These metaheuristics are often hybridized with local search to achieve state-of-the-art performance on challenging combinatorial optimization problems \citep[see, e.g.,][]{Falkenauer1996,Vidal2012,Mecler2021,Vidal2022}.
In a sense, hybrid GAs with local search extends the capabilities of ILS approaches, as they generate new starting points for the local search by \emph{recombining} existing solutions from a \emph{population} of high-quality local minima found in previous iterations. This way, the generation of new starting points is tightly connected to their success in the search history. 

Specific to clustering problems, \citet{pernkopf2005genetic} proposed a genetic-based EM called GAEM, which selects the number of model components using the minimum description length (MDL) criterion and exploits a single-point crossover \citep{michalewicz2013solve, fogel1997evolutionary}. For the minimum-sum-of-squares clustering problem (the model for which k-means is the usual choice), \citet{gribel2019hg} proposed a hybrid genetic algorithm called HG-means. This algorithm relies on a recombination operation built upon a bipartite matching problem and mutation steps to generate diverse and promising start points based on the information of the population. It achieved state-of-the-art results on a wide variety of benchmark instances compared to other popular clustering algorithms for the same model. Finally, in addition to genetic algorithms, other works have proposed split-and-merge techniques to escape from local minima. In Split-and-Merge EM (SMEM) \citep{ueda2000smem}, a pair of clusters is merged and another cluster is split in between two EM iterations. Stepwise SMEM (SSMEM) \citep{xian2004estimation} goes further and uses the SMEM strategy to estimate the optimal number of Gaussians in the mixture. Finally, Competitive EM \citep{zhang2004competitive} is another SMEM variant with different criteria to select which clusters to merge and split.

The aforementioned algorithms can lead to improved solutions; however, as shown in the experiments, these algorithms do not generalize well in a high-dimensional cases with few observations, resulting in over-fitted solutions. To circumvent this issue, we introduce \myblue{an optimization} algorithm using regularization techniques leading to better-conditioned matrices and better generalization as the outcome.

\section{A Hybrid Genetic EM with Regularization}
\label{section:methodology}

In this section, we introduce \myblue{an optimization} algorithm for ellipsoidal clustering. This algorithm is grounded on evolutionary computation and combines the general EM local search with regularization strategies. Evolutionary algorithms apply natural selection, crossover, and mutation principles to improve a population of solutions. In such algorithms, it is also essential to balance selection pressure and diversification to search for a wide spectrum of solutions.

The hybrid genetic search (HGS) we propose follows the same basic principles as \citet{gribel2019hg} and, therefore, combines the classical variation operators (crossover and mutation) with local improvement through EM. We use the regularization techniques presented in the previous section to avoid overfitting and bad conditioning of the covariance matrices. The general structure of the algorithm is summarized in \algorithmref{algorithm-general-structure}. The rest of this section details each component of the method: the solution representation and population initialization methods, the variation (crossover and mutation) operators for solution generation, the local search with a regularized EM algorithm, and the population-management methods.

\begin{algorithm}[htbp]
\caption{General structure of HGS}
\begin{algorithmic}[1]
\State Initialize population \Comment{$\triangleright$ \sectionref{subsection:initial}} 
\For{$T$ generations without improvement}
\State Select parents $p_1$ and $p_2$ \Comment{$\triangleright$  \sectionref{subsection:selection-crossover-mutation}}
\State Generate an offspring $\theta$ from $p_1$ and $p_2$ (crossover) \Comment{$\triangleright$ 
 \sectionref{subsection:selection-crossover-mutation}}
\State Generate an individual $\theta'$ by mutating $\theta$ (mutation) \Comment{$\triangleright$ 
 \sectionref{subsection:selection-crossover-mutation}}
\State Apply clustering local improvement on $\theta'$ \Comment{$\triangleright$  \sectionref{subsection:localsearch}}
\State Add $\theta'$ to the population
\If{population size exceeds maximum size $\Pi_{\text{max}}$} \Comment{$\triangleright$  \sectionref{subsection:management}}
\State Remove all clones in the population
\State Select survivors until the population reach $\Pi_{\text{min}}$ 
\EndIf
\EndFor
\State \Return best solution
\end{algorithmic}
\label{algorithm-general-structure}
\end{algorithm}
 
\subsection{Solution Representation and Population Initialization}
\label{subsection:initial}

Each solution $s$ is represented in the HGS as a set of three chromosomes:
\begin{itemize}[nosep]
\item $\Mu = \{\vec{\mu}_1, \dots, \vec{\mu}_k\}$ containing the $d$-dimensional coordinates of the $k$ clusters centers;
\item $\Sigma = \{\Sigma_1, ..., \Sigma_k\}$ defining the ${\rm I\!R}^{d \times d}$ covariance matrix of each cluster;
\item $\vec{\pi} = \{\pi_1, ..., \pi_k\}$ representing the mixture weights of each cluster.
\end{itemize}

To obtain an initial population containing diverse solutions, the algorithm creates $\Pi_{\text{max}}$ solutions by (i) randomly assigning the center of each cluster to random sample locations with uniform probability, (ii) initializing the covariance matrix of each cluster to the identity matrix and setting uniform initial mixture weights, and (iii) applying the local search algorithm (described in~\sectionref{subsection:localsearch}) from this starting point.

The fitness of each solution is then calculated as the log-likelihood of \equationref{equation-objective}:
\begin{align}
    \sum_{i=1}^{n} \log \left( \sum_{j=1}^{k} \pi_j \ \mathcal{N}(\vec{x}_i \ | \ \vec{\mu}_j, \Sigma_j) \right)
    \label{equation-objective}
\end{align}
\begin{align}
\mathcal{N}(\vec{x}_i \ | \ \vec{\mu}_j, \Sigma_j) = \frac{1}{\sqrt{\left(2 \pi\right)^d \det \Sigma_j}}\exp \left( -\frac{1}{2} \left( \vec{x}_i - \vec{\mu}_j \right)^\top \Sigma_{j}^{-1} \left( \vec{x}_i - \vec{\mu}_j \right) \right).
\label{multivariate-gaussian-distribution}
\end{align}

\subsection{Solution Generation by Crossover and Mutation}
\label{subsection:selection-crossover-mutation}

The HGS uses a binary-tournament selection to obtain two parents, $P_1$ and $P_2$. As illustrated in \figureref{binarytournament}, two solutions are randomly selected with uniform probability in the population during each binary tournament, and the best one is retained.

\definecolor{color_G}{HTML}{8fbdbd}
\definecolor{color_P}{HTML}{d28fd2}

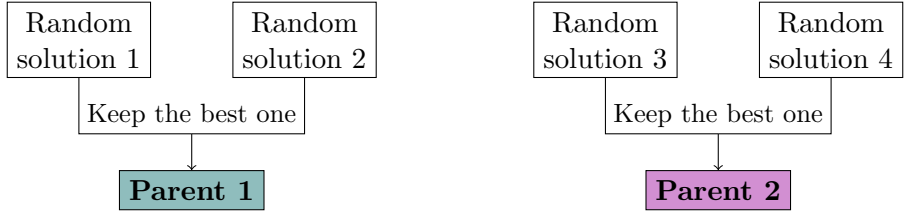
\begin{figure}[htbp]
\centering
\begin{tikzpicture}
\node[draw, align=center] at (-5,0) (s1) {Random\\solution 1};
\node[draw, align=center] at (-2,0) (s2) {Random\\solution 2};
\node[draw, align=center] at (+2,0) (s3) {Random\\solution 3};
\node[draw, align=center] at (+5,0) (s4) {Random\\solution 4};
\node[draw, align=center, fill=color_G] at (-3.5,-2) (p1) {\textbf{Parent 1}};
\node[draw, align=center, fill=color_P] at (+3.5,-2) (p2) {\textbf{Parent 2}};
\coordinate (p1f) at (-3.5,-1.25); 
\coordinate (p2f) at (+3.5,-1.25); 
\draw (s1) |- (p1f);
\draw (s2) |- (p1f);
\draw (s3) |- (p2f);
\draw (s4) |- (p2f);
\draw[->] (p1f) -- node [above=0.5em]{\small Keep the best one} (p1);
\draw[->] (p2f) -- node [above=0.5em]{\small Keep the best one} (p2);
\end{tikzpicture}
\caption{Diagram representing the workflow of the binary tournament}
\label{binarytournament}
\end{figure}

During the crossover step, the algorithm solves a matching problem in a bipartite graph $G = (V_1,V_2,E)$, in which each vertex $v \in V_1$ (respectively $V_2$) stands for a cluster in $P_1$ (respectively,~$P_2$), therefore $|V_1| = |V_2| = k$. Each edge $(i,j) \in E$ for $i \in V_1$ and $j \in V_2$ represents the possibility of associating the $i^\text{th}$ cluster of $P_1$ with the $j^\text{th}$ cluster of $P_2$ for an associated cost:

\begin{align}
 c_{ij} = \frac{\sqrt{\left(\vec{\mu}_i - \vec{\mu}_j\right)^{\top} \Sigma_{j}^{-1} \left(\vec{\mu}_i - \vec{\mu}_j\right)} + \sqrt{\left(\vec{\mu}_j - \vec{\mu}_i\right)^{\top} \Sigma_{i}^{-1} \left(\vec{\mu}_j - \vec{\mu}_i\right)}}{2}.
 \label{equation-hungarian}
\end{align}

\begin{figure}[htbp]
\centering
\subfigure[Parent 1]{
\includegraphics[width=0.22\textwidth]{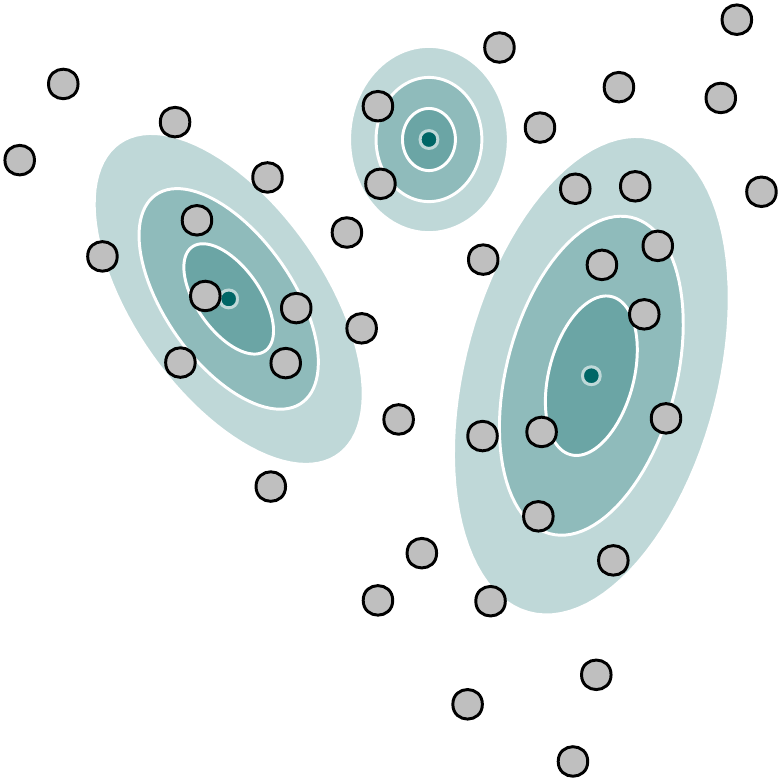}
\label{}
}
\hspace{\fill}
\subfigure[Parent 2]{
\includegraphics[width=0.22\textwidth]{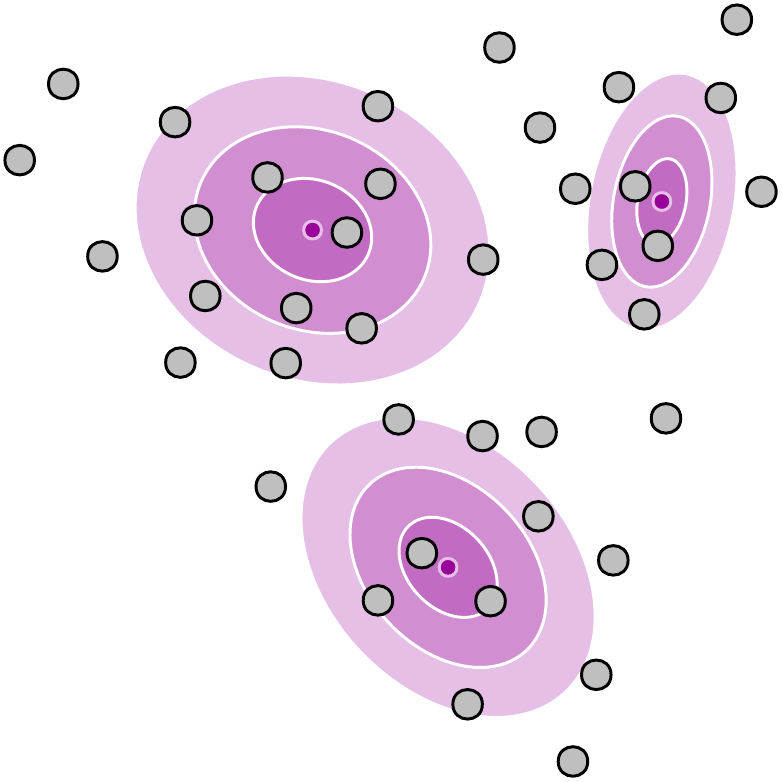}
\label{}
} 
\hspace{\fill}
\subfigure[Pairwise centers]{
\includegraphics[width=0.22\textwidth]{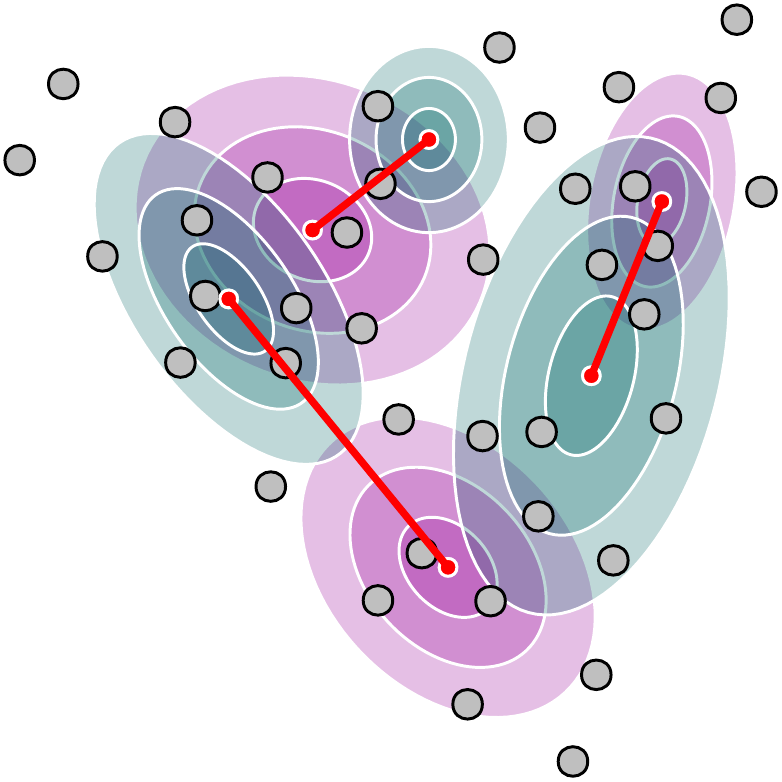}
\label{}
}
\hspace{\fill}
\subfigure[Offspring]{
\includegraphics[width=0.22\textwidth]{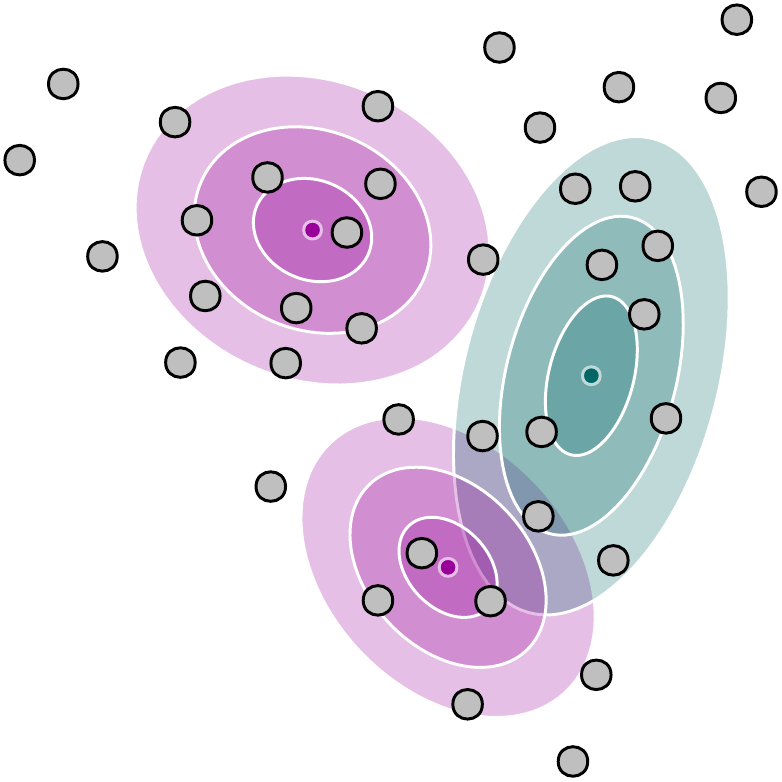}
\label{}
}
\caption{Crossover operator: (a) and (b) represent the parents; (c) solution of the matching step; (d) offspring obtained after retaining one cluster for each matched pair}
\label{figure-crossover}
\end{figure}

Finding a minimum-cost bipartite matching in this graph gives us a better association between the centers of the two parents. We use the Hungarian Algorithm \citep{kuhn1955hungarian} to obtain this matching. Then, for each pair of clusters (i.e., edge) of the bipartite matching solution, we randomly retain one of the two clusters in the child along with its covariance matrix. At the same time, the mixture coefficient of this cluster is defined as the average from both matched clusters. This process effectively permits inheriting centers from both parents while keeping a good distribution of centers. \figureref{figure-crossover} illustrates the crossover operator.

\begin{figure}[htbp]
\centering
\subfigure[Solution]{
\includegraphics[width=0.22\textwidth]{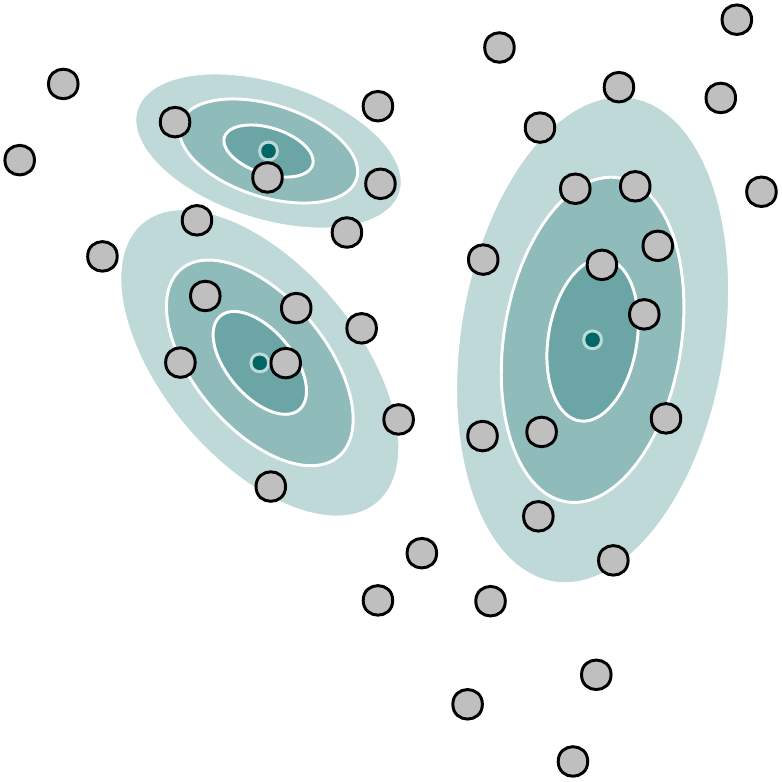}
\label{}
}
\hspace{\fill}
\subfigure[Random points]{
\includegraphics[width=0.22\textwidth]{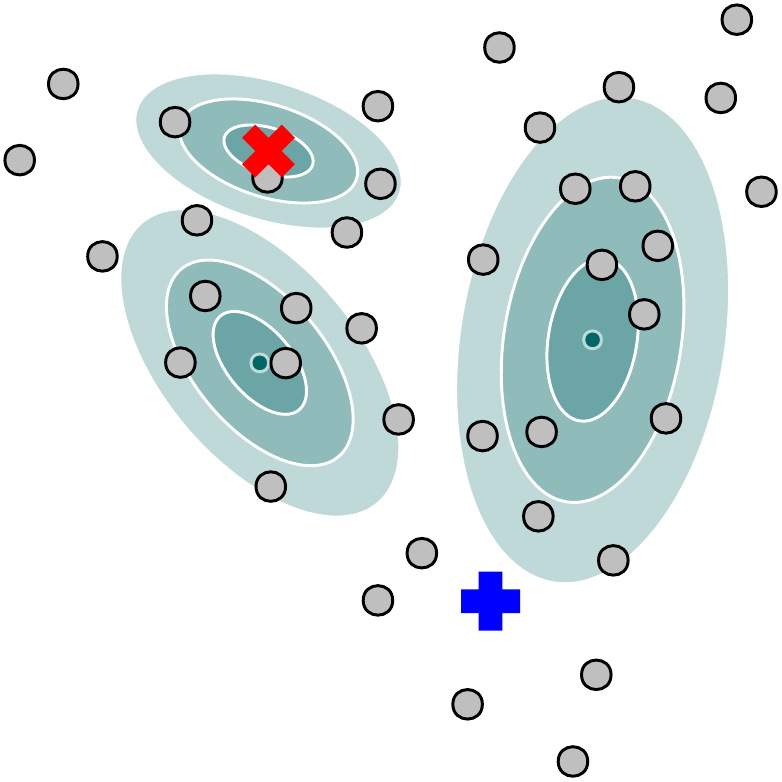}
\label{}
} 
\hspace{\fill}
\subfigure[Reallocation]{
\includegraphics[width=0.22\textwidth]{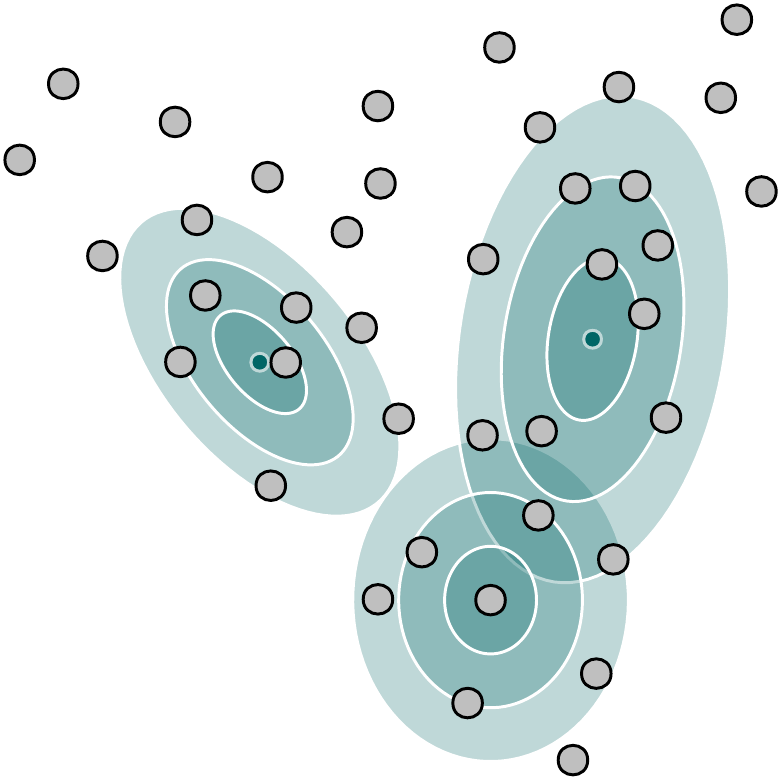}
\label{}
}
\hspace{\fill}
\subfigure[Solution improvement]{
\includegraphics[width=0.22\textwidth]{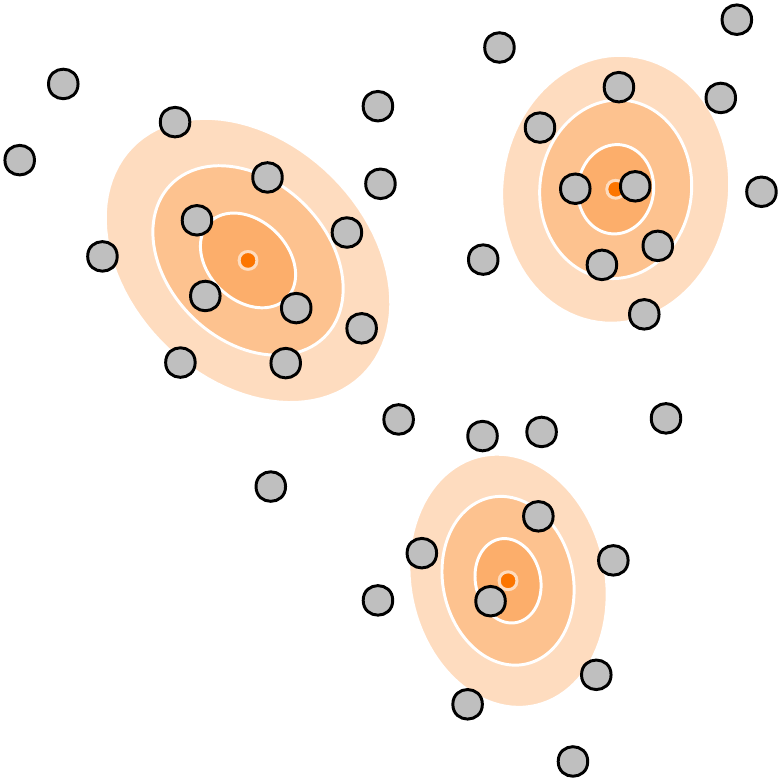}
\label{}
}
\caption{Mutation operator: (a) current solution; (b) selection of a random cluster (red $\times$ mark) and random data sample (blue $+$ mark); (c) reallocation of the cluster; (d) resulting solution after local search}
\label{figure:mutation}
\end{figure}

The solution generated by the crossover operator passes through a mutation step, which is designed to diversify the search. The mutation operator randomly selects a cluster and a data sample according to a uniform probability distribution. The cluster is reallocated to the position of the chosen data sample, and the covariance matrix is re-initialized as the average of the other clusters' covariance matrices. This maintains the relative scale of the clusters' parameters and effectively prevents situations in which one specific cluster would progressively lose importance in the mixture and ultimately disappear. 
Figure \ref{figure:mutation} illustrates the mutation operator as well as the final solution obtained after applying the local search (described in \sectionref{subsection:localsearch}) on it.

\subsection{Local Search with a Regularized EM}
\label{subsection:localsearch}

The solution produced by the crossover and mutation operators serves as the starting point for a local search using a regularized variant of the EM algorithm for GMM. Algorithm \ref{gmm-generalstructure} describes the general structure of this algorithm.
  
\begin{algorithm}
\caption{EM-GMM Clustering}
\begin{algorithmic}[1]
\State Initialize solution 
\While{the stopping criterion has not been met}
\State Compute $\gamma$ ownership weights 
\State Recompute the mixing weights $\pi$ 
\State Recompute the clusters' centers $\mu$
\State Recompute the covariance matrices $\Sigma$
\EndWhile
\State \Return solution
\end{algorithmic}
\label{gmm-generalstructure}
\end{algorithm}

The specialization of the EM algorithm for GMM clustering is described by \cite{Bishop2006} and \cite{friedman2001elements}. The EM alternates between expectation and maximization steps; it optimizes the mixture's parameters along with latent variables representing the ownership weight of each sample $\in X$, i.e., the probability of attribution of this point to each cluster, also known as responsibility. The \emph{expectation} step is described in Equation (\ref{4ownership}). Given the current solution ($\pi$,~$\vec{\mu}$,~$\Sigma$), this step computes the estimated ownership weights of each sample belonging to each cluster:

\begin{align}
\gamma_{i, j} = \frac{\pi_j \ \mathcal{N}(\vec{x}_i \ | \ \vec{\mu}_j, \Sigma_j)}{\sum_{c=1}^k \pi_c \ \mathcal{N}(\vec{x}_i \ | \ \vec{\mu}_c, \Sigma_c)}.
\label{4ownership}
\end{align}

In contrast, the \emph{maximization} step re-optimizes the parameters of the EM-GMM ($\pi$, $\vec{\mu}$ and $\Sigma$) based on the values of the latent variables, as depicted in Equations (\ref{4max1}--\ref{4max3}):
\begin{align}
\pi_j &= \sum_{i=1}^{n} \gamma_{i, j} \left/ \sum_{j=1}^{k} \sum_{i=1}^{n} \gamma_{i, j} \right. \label{4max1} \\
\vec{\mu}_j &= \sum_{i=1}^{n} \gamma_{i, j} \vec{x}_i \left/ \sum_{i=1}^{n} \gamma_{i, j} \right. \label{4max2} \\
\Sigma_{j} &=  \sum_{i=1}^n \gamma_{i, j} \left(\vec{x}_i - \vec{\mu}_j \right) \left(\vec{x}_i - \vec{\mu}_j \right)^\top \left/ \sum_{i=1}^{n} \gamma_{i, j} \right. . \label{4max3}
\end{align}

\begin{figure}[htbp]
\centering
\subfigure[Iteration 1]{
\includegraphics[width=0.22\textwidth]{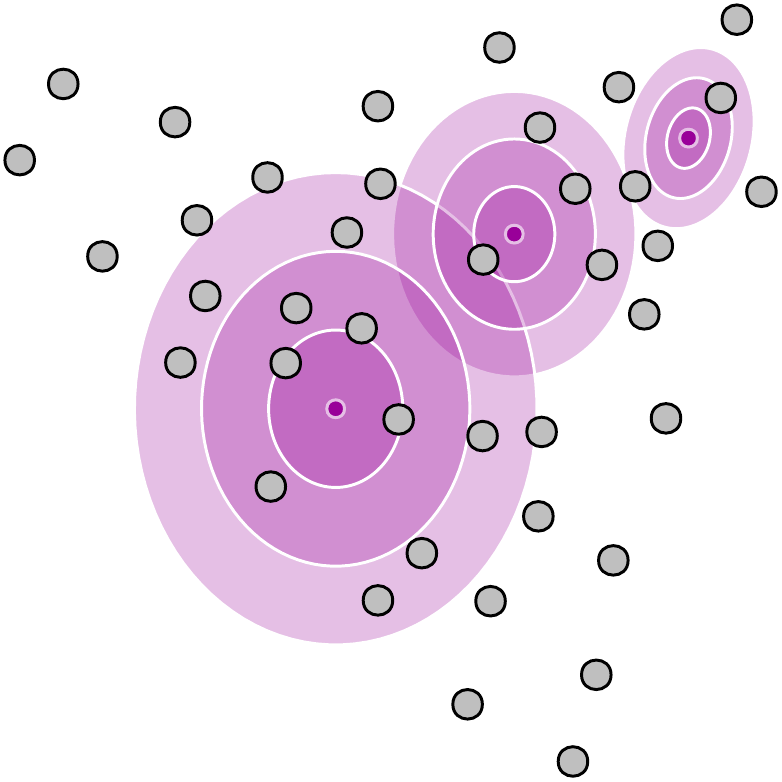}
\label{}
}
\hspace{\fill}
\subfigure[Iteration 9]{
\includegraphics[width=0.22\textwidth]{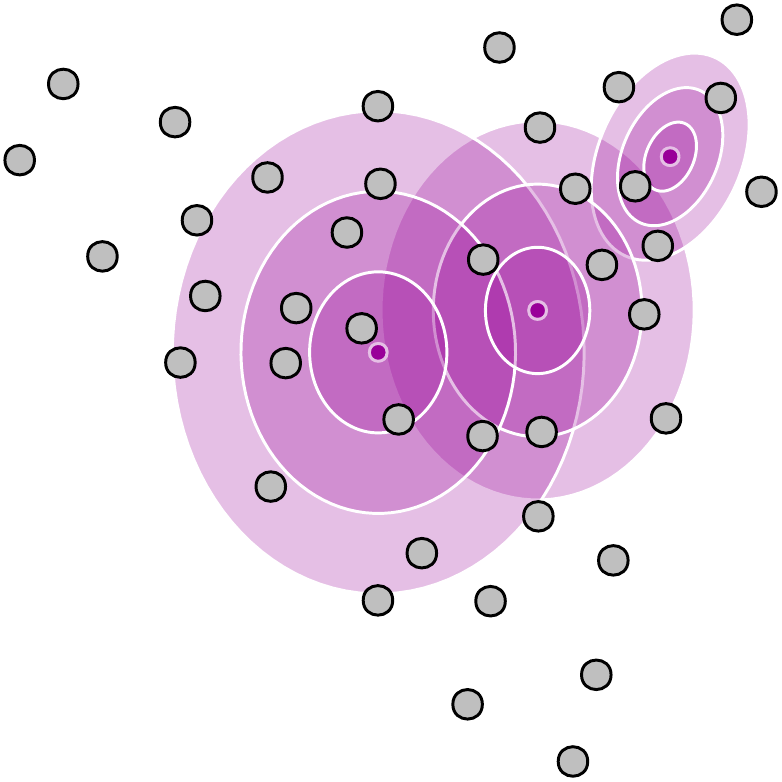}
\label{}
} 
\hspace{\fill}
\subfigure[Iteration 14]{
\includegraphics[width=0.22\textwidth]{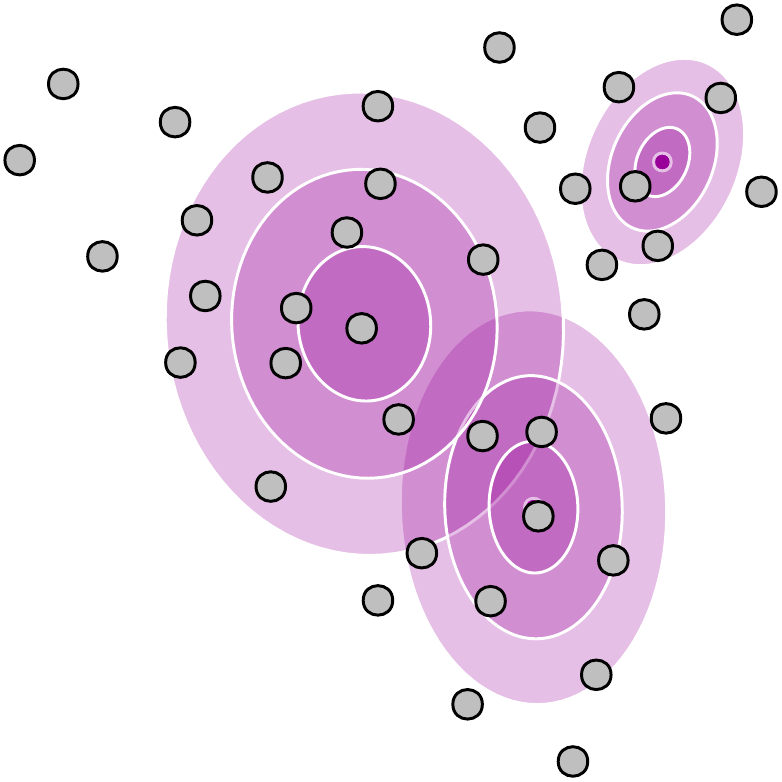}
\label{}
}
\hspace{\fill}
\subfigure[Iteration 32]{
\includegraphics[width=0.22\textwidth]{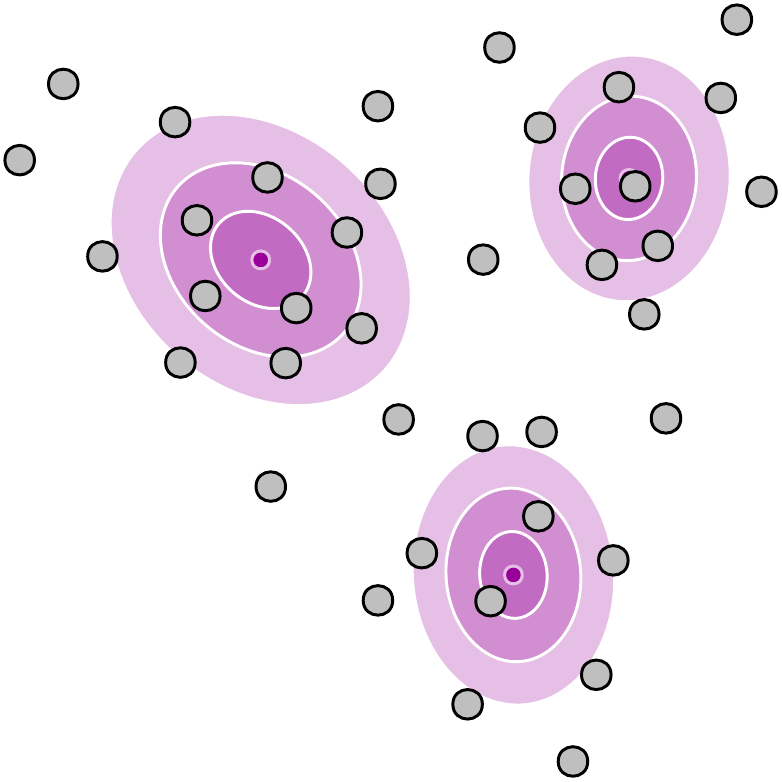}
\label{}
}
\caption{Successive solutions in the EM-GMM}
\label{figure:improvement}
\end{figure}

Each of these two steps is guaranteed to improve the likelihood, as it concentrates on re-optimizing a subset of the parameters. The EM algorithm ``iterates'' by alternating between these two steps until no more improvement is achievable. An example of the progression of this algorithm is depicted in \figureref{figure:improvement}.\\

\noindent
\textbf{Regularization in EM.} As previously discussed, the classical EM-GMM suffers from poor conditioning of the covariance matrices. To circumvent this issue, we use regularization techniques applied at each iteration of the algorithm immediately after estimating the empirical covariance matrix in the maximization step. We consider three alternative regularization techniques that can be efficiently implemented. These approaches impact how parameter $\delta$ of \equationref{eq:covariance-shrunk} is selected.

\begin{enumerate}
    \item The first regularization method, called Shrunk methodology, consists of fixing the parameter $\delta_{\text{Shrunk}} = 0.1$. This is the default approach adopted in scikit-learn \citep{scikitlearn}.
    \item The second regularization method, called Ledoit-Wolf Shrinkage (LW), attempts to minimize the mean squared error between the estimated and the real covariance matrix. The resulting matrix can be calculated in closed form, as seen in \citet{ledoit2004well}.
    \item The last regularization method, called Oracle Approximating Shrinkage (OAS -- \citealt{chen2010shrinkage}) assumes that the data came from a Gaussian distribution. According to that study, the shrinkage coefficient $\delta$ reduces further the mean squared error compared to the LW regularization. Equation~(\ref{oas-formula}) is the formula that computes the $\delta$ in OAS:
\begin{align}
\delta_{\text{OAS}} = \min \left(1, \frac{\left(1-2/d\right) \Tr\left(\Sigma^2\right) + \Tr^2\left(\Sigma\right)}{\left(n + 1 -2/d\right) \left[ \Tr\left(\Sigma^2\right) + \Tr^2\left(\Sigma \right)/d \right]}\right).
\label{oas-formula}
\end{align}
\end{enumerate}
 
\begin{figure}[htbp]
\centering
\subfigure[256 samples]{\includegraphics[align=c,width=0.25\textwidth]{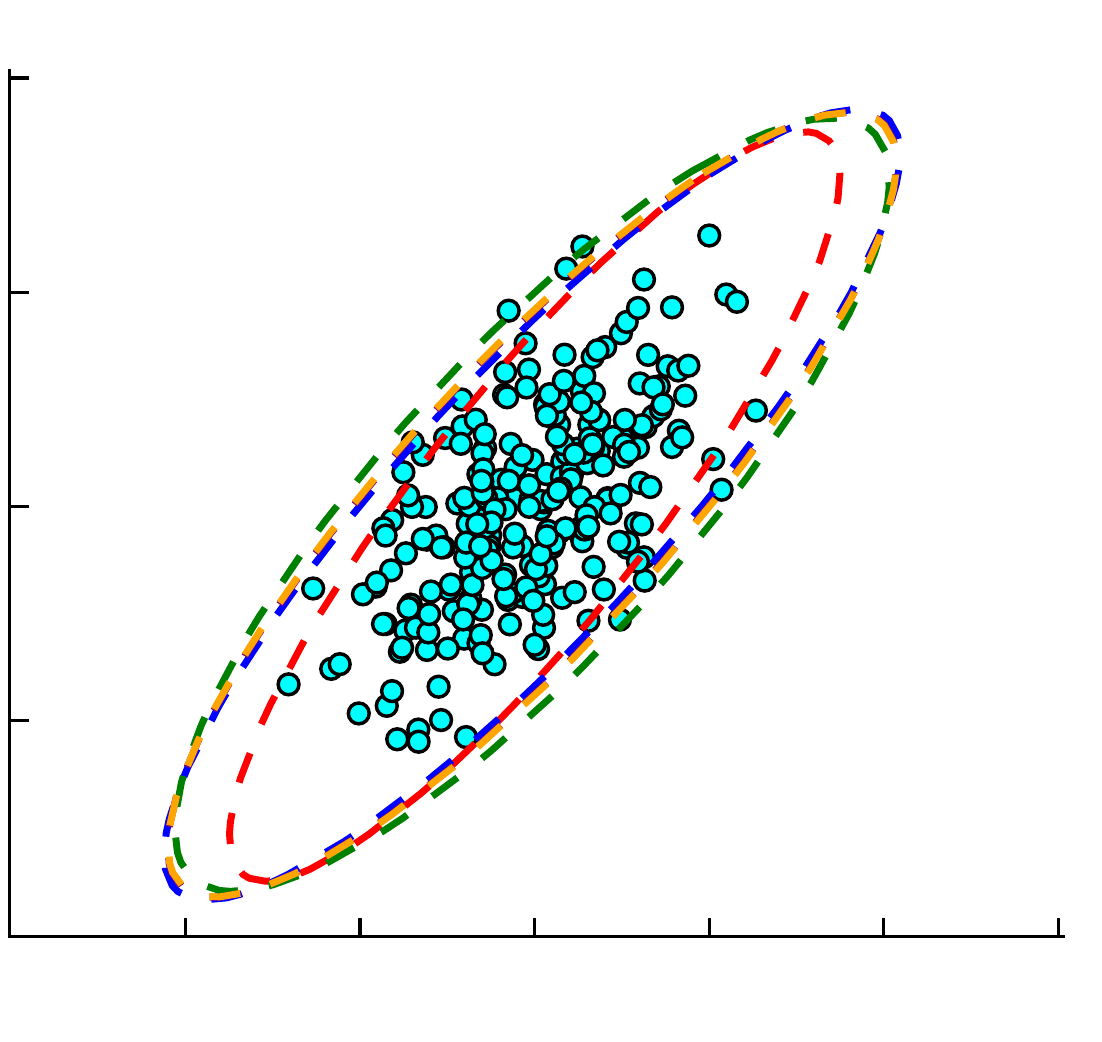}}
\subfigure[8 samples]{\includegraphics[align=c,width=0.25\textwidth]{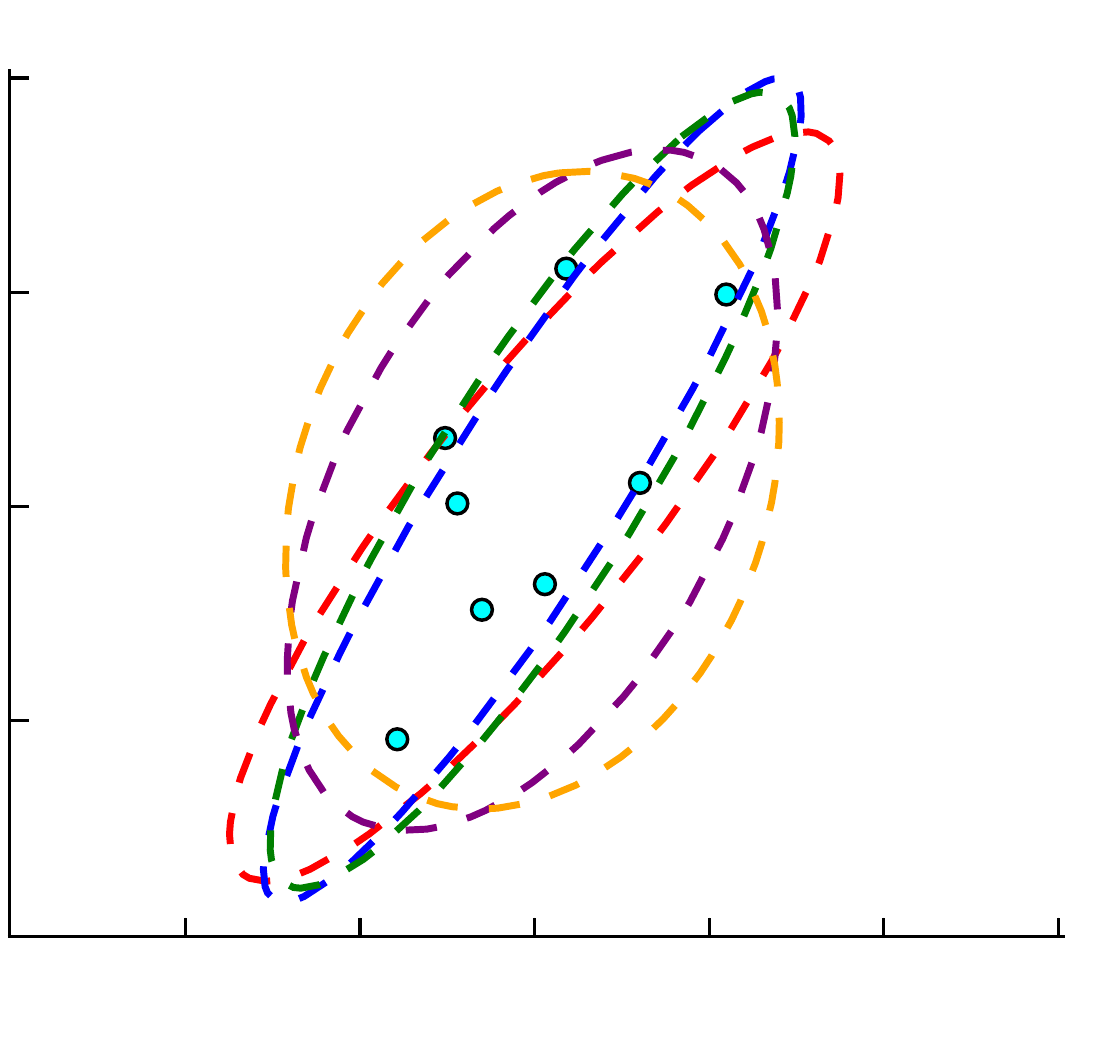}}
\subfigure[4 samples]{\includegraphics[align=c,width=0.25\textwidth]{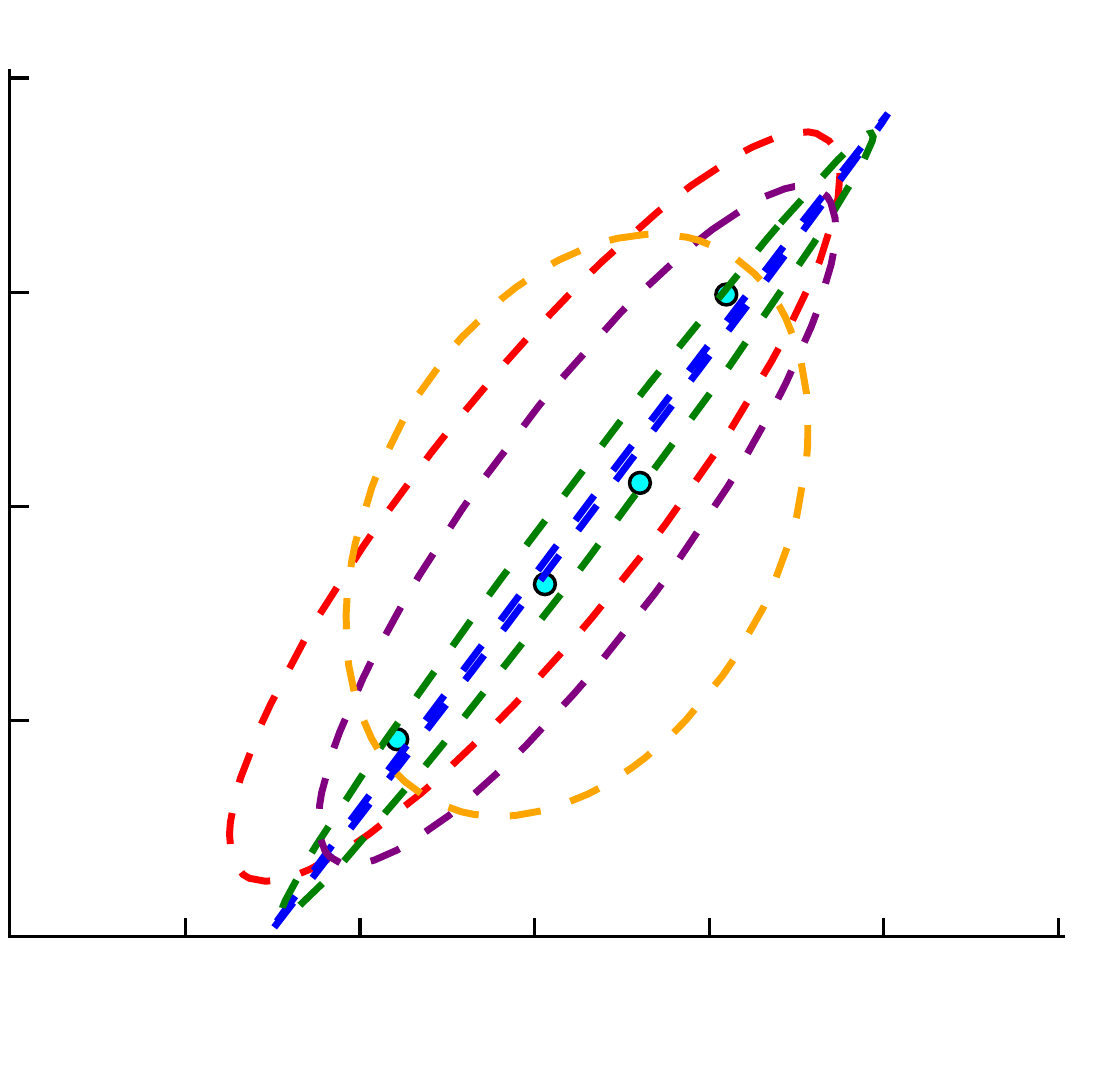}}
\subfigure{\includegraphics[align=c,width=0.20\textwidth]{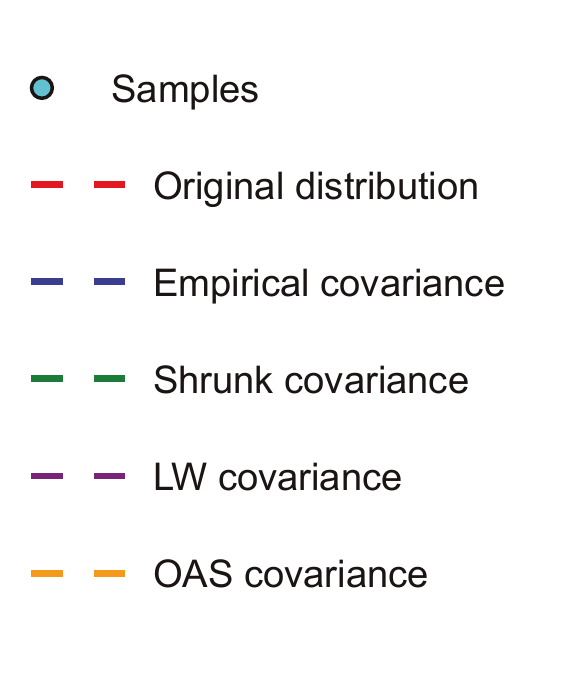}}
\caption{Behavior of the empirical and the regularized covariance estimations}
\label{fig:shrinkage}
\end{figure}

\figureref{fig:shrinkage} compares the results of the different covariance estimation methods. As visible in this figure, the impact of regularization is more marked in situations where few data points are available to estimate the covariance matrices. In those cases, the regularization strategies provide better-conditioned matrices.

\subsection{Population Management}
\label{subsection:management}

Finally, whenever the population exceeds the maximum size of $\Pi_{\text{max}}$, HGS performs a survivor selection mechanism to retain only $\Pi_{\text{min}}$ solutions. Survivor selection ensures a gradual selection pressure towards higher-quality solutions. To preserve diversity, we implement a clone elimination mechanism, which detects individuals with the same objective value and prioritizes removing one of them. Once no such clone exists anymore and if the population still exceeds $\Pi_{\text{min}}$, the algorithm proceeds by removing the solutions with the smallest objective value (i.e., smallest likelihood) until it reaches the desired population size.

\section{Computational Experiments}
\label{section:experiments}

We conduct extensive computational experiments with two main goals: (i) to measure the impact of regularization techniques on cluster recovery and (ii) to assess the importance of more sophisticated \myblue{optimization} algorithms and their interaction with the regularization strategies. To that end, we implemented the proposed methods in \myblue{Julia v$1.8.5$} \citep{julialang} and conducted our experiments on an \myblue{Intel(R) Xeon(R) Platinum 8375C CPU @~2.90~GHz}.
All the source code, scripts, and data needed to reproduce these experiments are openly accessible at \myblue{\url{http://www.github.com/raphasampaio/RegularizationAndOptimizationInModelBasedClustering.jl}}.

\subsection{Datasets and Experimental Setup}
\label{sec:Exp-Data}

\begin{figure}[htbp]
\centering
\subfigure[$c=-0.26$]{
\boxed{\includegraphics[width=0.2\textwidth]{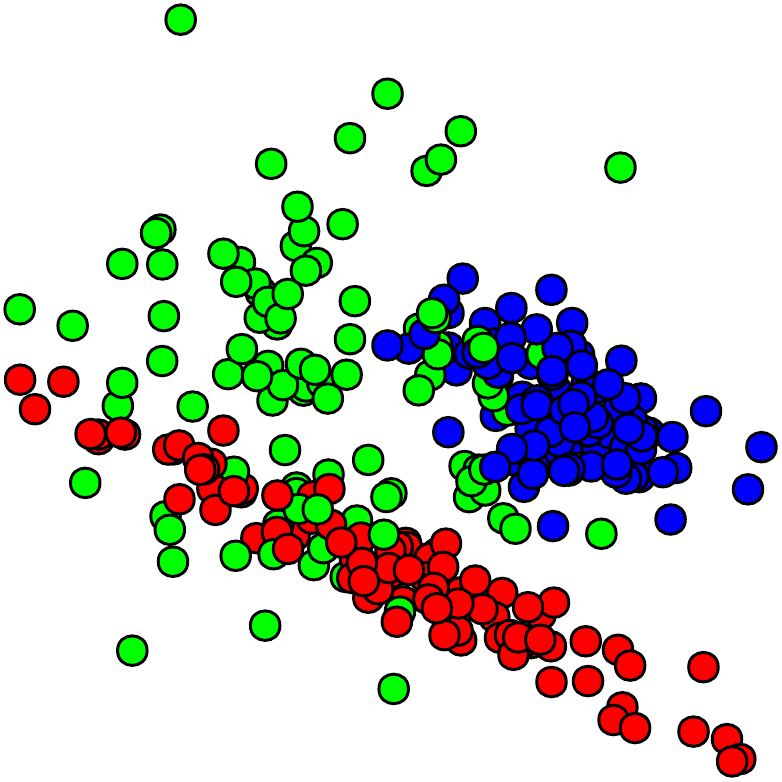}}
\label{}
}
\hspace{\fill}
\subfigure[$c=-0.10$]{
\boxed{\includegraphics[width=0.2\textwidth]{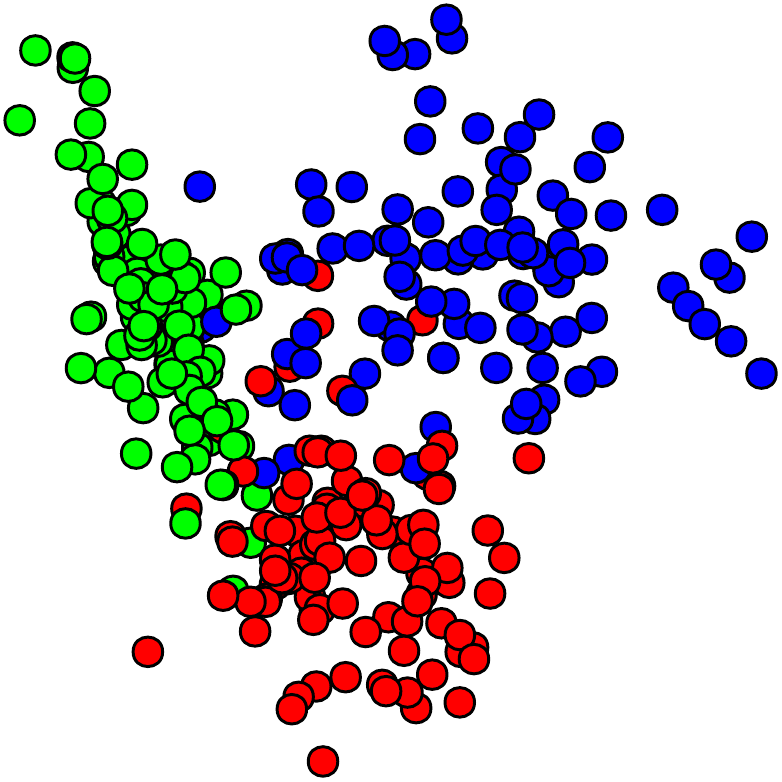}}
\label{}
} 
\hspace{\fill}
\subfigure[$c=0.01$]{
\boxed{\includegraphics[width=0.2\textwidth]{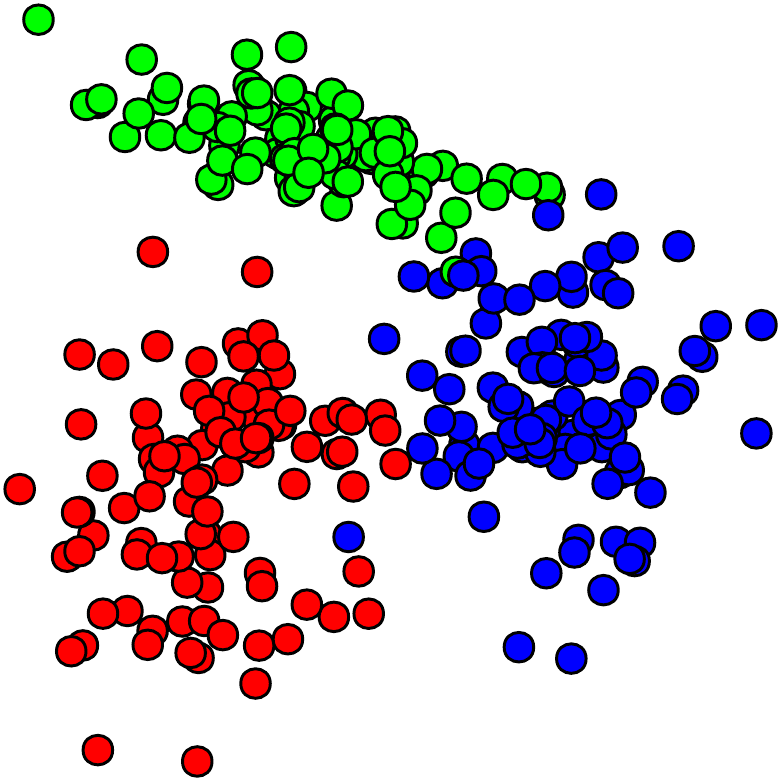}}
\label{}
}
\hspace{\fill}
\subfigure[$c=0.21$]{
\boxed{\includegraphics[width=0.2\textwidth]{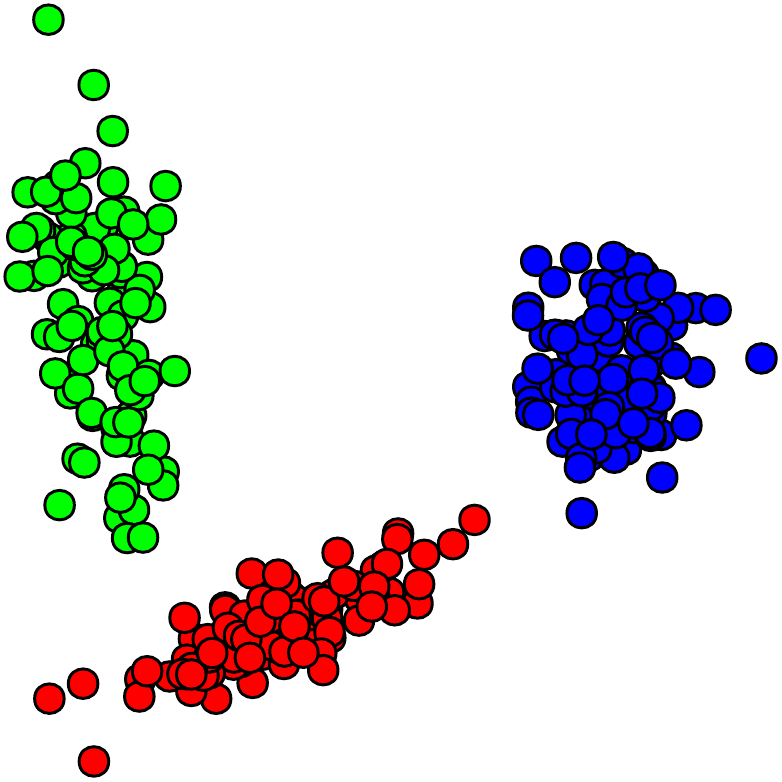}}
\label{}
}
\caption{Examples of instances generated with different levels of the separability index $c$}
\label{fig:separability}
\end{figure}

We generated the benchmark instances using the \textit{ClusterGeneration} package \citep{clusterGeneration} in~$R$ programming language. The difficulty of the clustering task for these datasets is governed by the Separation Index \citep{qiu2006generation,qiu2006separation}. The higher this index, the more separated the clusters are (see~\figureref{fig:separability}). We therefore generated datasets with different values of the separability index ($c \in \{-0.26, -0.10, 0.01, 0.21\}$), different number of features ($d \in \{2, 5, 10, 20, 30, 40\}$) and different number of clusters ($k \in \{3, 10, 20\}$). For each configuration, described as a tuple $(c,d,k)$, we fixed the dataset size ($n = 100 \cdot k$) and generated $20$ random datasets. In all datasets, the covariance matrix's eigenvalues describing each cluster's shape were randomly sampled from a uniform distribution over the range $\left[1, 200\right]$. \tableref{table:configurations} summarizes the main factors considered in our analyses and their possible levels, i.e., the parameters governing the structure of the instances.

\begin{table}[htbp]
\centering
\scalebox{0.95}
{
\begin{tabular}{l@{\hspace{1cm}}l}
\toprule
\textbf{Factors} &  \textbf{Levels}\\
\midrule
Separability ($c$)  & $\{-0.26, -0.10, 0.01, 0.21\}$        \\
Number of Features ($d$)  & $\{2, 5, 10, 20, 30, 40\}$      \\
Number of Clusters ($k$) & $\{3, 10, 20\}$                  \\
\bottomrule
\end{tabular}
}
\caption{Main factors considered in our experiments}
\label{table:configurations}
\end{table}


We adopt three similarity measures to compare cluster assignments: the Adjusted Rand Index (ARI) \citep{hubert1985comparing}, the Normalized Mutual Information (NMI) \citep{vinh2009information}, and the Centroid index (CI) \citep{franti2014centroid}.

The ARI is the corrected version of the Rand index (RI) \citep{rand1971objective, vinh2010information}, a widely used measure for clustering analysis \citep{steinley2004properties}. The ARI establishes a baseline using the expected similarity of all pairwise comparisons between the classifications. The ARI score belongs to the range $[-1, 1]$, where $1$ corresponds to perfect assignments identical to the true values, and $0$ would stand for random uniform cluster assignments.

The NMI is an information-theoretic measure that gauges the similarity between two clusterings by considering the entire data as a single set. It compares the information within each clustering with the combined total information from both clusterings. Essentially, NMI assesses how much information one clustering shares with the other. The NMI score belongs to the range $[0, 1]$, where a score close to $1$ indicates that the two clusterings are almost identical, and a score close to $0$ implies minor to no mutual information between the two clusterings.

The CI is a set-matching-based measure. Instead of considering pairwise relations or mutual information, CI focuses on how centroids (or representative elements) of clusters in one clustering relate to those in the other. By examining the alignment and proximity of these centroids, CI offers a geometric perspective on clustering similarity. The CI score belongs to the range $[0, k-1]$, where two solutions with an equal number of clusters are said to have the same cluster-level structure if each prototype from one clustering is mapped precisely once to a prototype in the other, resulting in a CI score equals to $0$. If this exact mapping does not occur, any unmatched or ``orphan'' prototype signifies a cluster absent or represented differently in the other clusterization.

We conduct our analyses in several steps, using the standard GMM algorithm (i.e., ``Local Search'' with ``Empirical'' covariance estimation) as our fundamental baseline and assessing the performance using ARI. First, we will vary the regularization techniques within the GMM to evaluate their impact (\sectionref{sec:Regularization-GMM}). Next, we will combine the use of different regularization techniques with more sophisticated search algorithms to observe their interactions (\sectionref{sec:Regularization-Global}). In these two analyses, we use the synthetic datasets described in \sectionref{sec:Exp-Data} to measure the impact of factors related to the characteristics of the instances. Next, \sectionref{sec:K-Means} extends our analyses to include additional methods (k-means), evaluations based on NMI and CI, and practical considerations (computational time). Finally, \sectionref{sec:UCI} reports results on datasets issued from the UCI, whose characteristics are more diverse but less controllable.

\subsection{Impact of Regularization}
\label{sec:Regularization-GMM}

In this section, we study the impact of covariance estimation with regularization techniques. When the square of the number of features, $d^2$, grows large compared to the number of samples $n$, the sample covariance matrices become ill-conditioned and represent poor estimators of the sample distribution. The use of regularization techniques can help overcome these issues. To estimate the impact of regularization, we first compare the performance of EM-GMM (for short GMM), without covariance matrix regularization with its counterpart with different regularization techniques (Shrunk, OAS, and LW -- defined in the previous section).

\begin{figure}[htbp]
\centering
\includegraphics[width=0.9\textwidth]{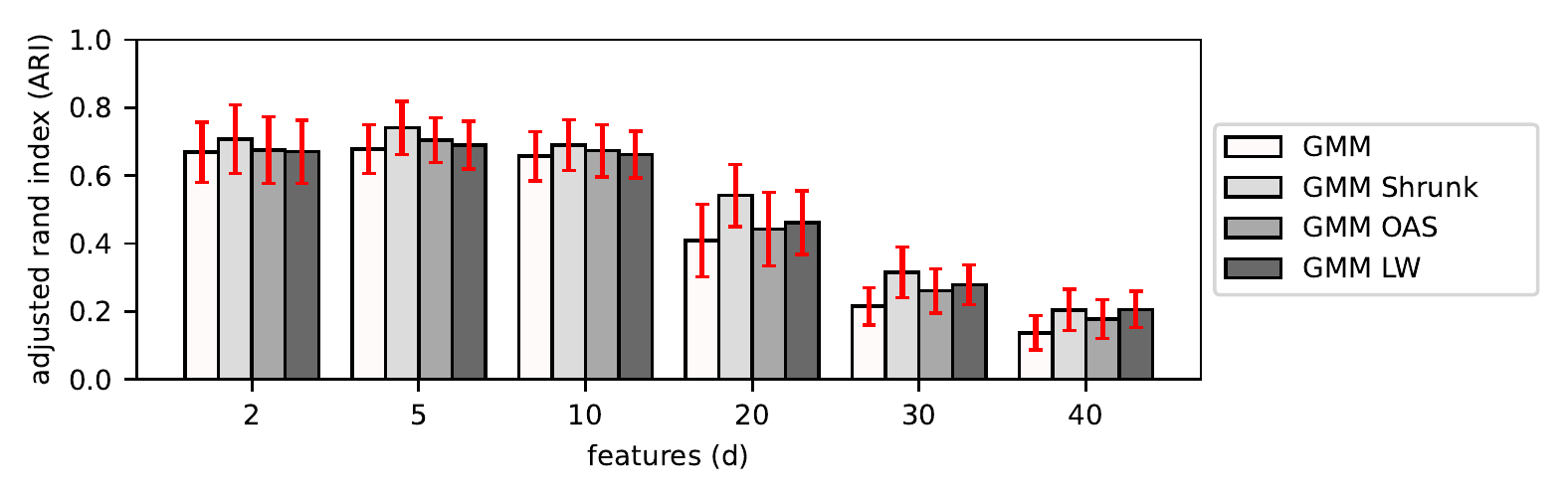}
\caption{Performance of the GMM and its regularized variants, for datasets with $k = 10$ and $c = 0.01$}
\label{fig:regularization}
\end{figure}

Therefore, \figureref{fig:regularization} compares the clustering performance of the different approaches in terms of~ARI. To measure the impact of the number of features, we report the results for varying values of~$d$ while keeping the parameters governing the number of clusters and separability index fixed to $k = 10$ and $c = 0.01$.
For each method and value of $d$, we average the results obtained from applying the method in all 20 datasets and represent it as a barplot with an additional interval representing the standard deviation. In all cases, results from regularized versions of GMM are equal to or better than standard GMM. The cluster-recovery performance still seems to degrade as the number of features grows, as a consequence of possible overfitting, but the decay is slower for methods that include covariance matrix regularization.

\begin{figure}[htbp]
\centering
\includegraphics[width=0.85\textwidth]{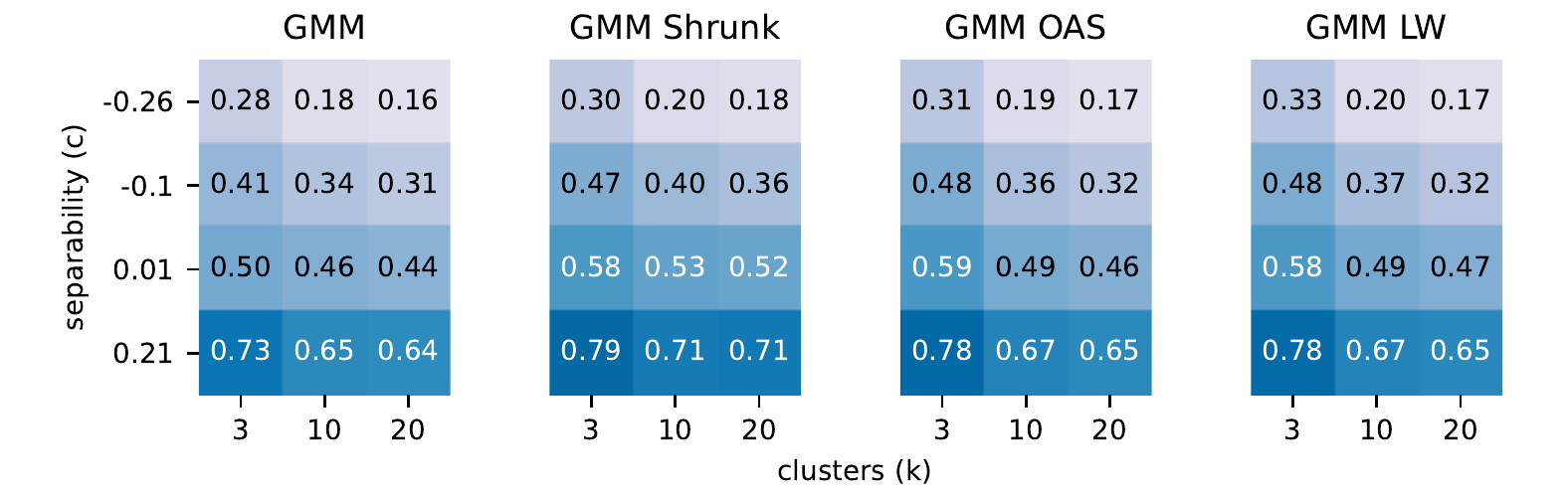}
\caption{Average of the ARI performance of GMM and its regularized variants over the $d = \{2, 5, 10, 20, 30, 40\}$ datasets for different separability indices and number of clusters}
\label{fig:heatmap3}
\end{figure}

In \figureref{fig:heatmap3}, we further analyze the impact of two additional factors: the dataset's separability index~$c$ and the number of clusters~$k$.
Our results are represented as four heatmaps, each corresponding to GMM algorithms with different covariance estimations: GMM (empirical estimation), GMM Shrunk, GMM OAS, and GMM LW. The rows correspond to different values of the separability index, and the columns correspond to different numbers of clusters. Each cell is the average over 120 datasets (20 different datasets for each number of features: $d = \{2, 5, 10, 20, 30, 40\}$).

As seen in \figureref{fig:heatmap3}, the Shrunk methodology outperformed OAS, LW, and no regularization when the number of clusters increases. The significance of these performance differences is confirmed by pairwise Wilcoxon tests between the results of GMM Shrunk and those of other methods at a 0.05\% significance level, presented in~\tableref{tab:wilcoxon1}.

\begin{table}[htbp]
\centering
\scalebox{0.9}
{
\begin{tabular}{l@{\hspace{1cm}}l}
\toprule
\textbf{Pair of methods} & \textbf{p-value} \\
\midrule
GMM Shrunk -- GMM & $2.17\times 10^{-127}$ \\
GMM Shrunk -- GMM OAS & $7.61\times 10^{-57}$ \\
GMM Shrunk -- GMM Ledoitwolf & $1.46\times 10^{-42}$ \\
\bottomrule
\end{tabular}
}
\caption{Impact of regularization on GMM: Pairwise Wilcoxon tests}
\label{tab:wilcoxon1}
\end{table}

These differences are likely due to the fact that OAS and LW have a larger impact on the shape matrix than the Shrunk approach. Regularization methods that estimate the data shape as too spherical reduce the possible space of clustering solutions. In contrast, Shrunk ensures that the shape matrix is well conditioned but has a more limited impact on it. Due to its good results and simplicity, this regularization approach will be used in the remainder of our experiments.

\subsection{Combining Regularization and \myblue{Optimization}}
\label{sec:Regularization-Global}

In this section, we evaluate whether the Shrunk regularization combined with more sophisticated search strategies can achieve even better clustering performance.
We consider, to that end, three search strategies.
The first is a simple Multi Start (GMM MS) approach, which repeats the GMM from~$n_\textsc{it}$ starting points and retains the solution with maximum likelihood. The second is a Random Swap (GMM RS) method similar to \citet{zhao2012random} and \citet{franti2018efficiency}. This method iteratively applies the GMM to find a local minimum in terms of likelihood and then randomly relocates a cluster position to generate a new starting point for the GMM. The third and final approach is the Hybrid Genetic Search (GMM HG) described in \sectionref{section:methodology}, which generates new starting points for the GMM by recombination of previous solutions.
We set a maximum of $n_\textsc{it} = 100$ iterations without improvement for each meta-heuristic, where an iteration refers to one application of GMM from a starting point (obtained randomly in the case of MS, from a relocation process in RS, or the crossover and mutation operator in HG).

\begin{figure}[htbp]
\centering
\includegraphics[width=0.9\textwidth]{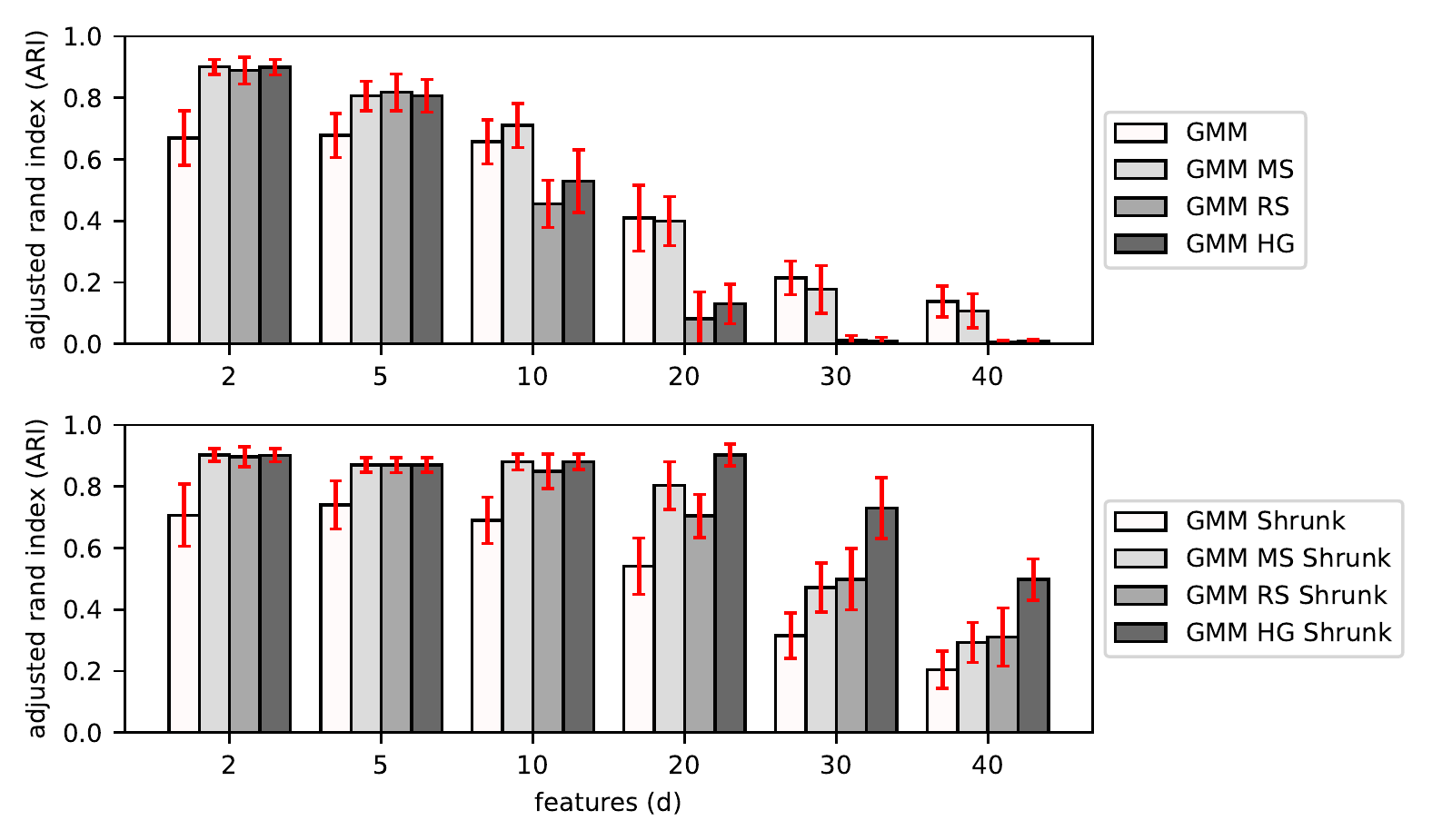}
\caption{Average and standard deviation of the ARI obtained by GMM and its \myblue{optimization} search strategies variations over the $k = 10$ and $c = 0.01$ datasets for different numbers of features, with and without regularization}
\label{fig:optimization5}
\end{figure}

\figureref{fig:optimization5} compares the ARI of the classical GMM with those achieved by the three improved search strategies, without and with Shrunk regularization. We focus this experiment on datasets for which the number of clusters and separability index are fixed to $k = 10$ and $c = 0.01$ while varying the values of $d$. For each method and value of $d$, we average the results and represent it as a barplot with an additional interval representing the standard deviation. 

The results presented in \figureref{fig:optimization5} lead to a rich set of observations. As it appears, the use of more sophisticated search approaches such as MS, RS, or HG leads to a performance deterioration when no regularization mechanism is used, especially for datasets with a large number of features. With the Shrunk regularization, however, using more sophisticated search approaches (HG, especially) leads to remarkable improvements in clustering performance. This connection between regularization and search intensity is a known phenomenon: without any proper control over the number of parameters, \myblue{superior solutions} will tend to overfit the assignment of samples to clusters and covariance matrix estimation. This effect is due to the number of parameters estimated when doing ellipsoidal clustering, which is $\bigO{kd^2}$ compared to $\bigO{kd}$ for classical k-means methods.

\begin{figure}[htbp]
\centering
\includegraphics[width=0.85\textwidth]{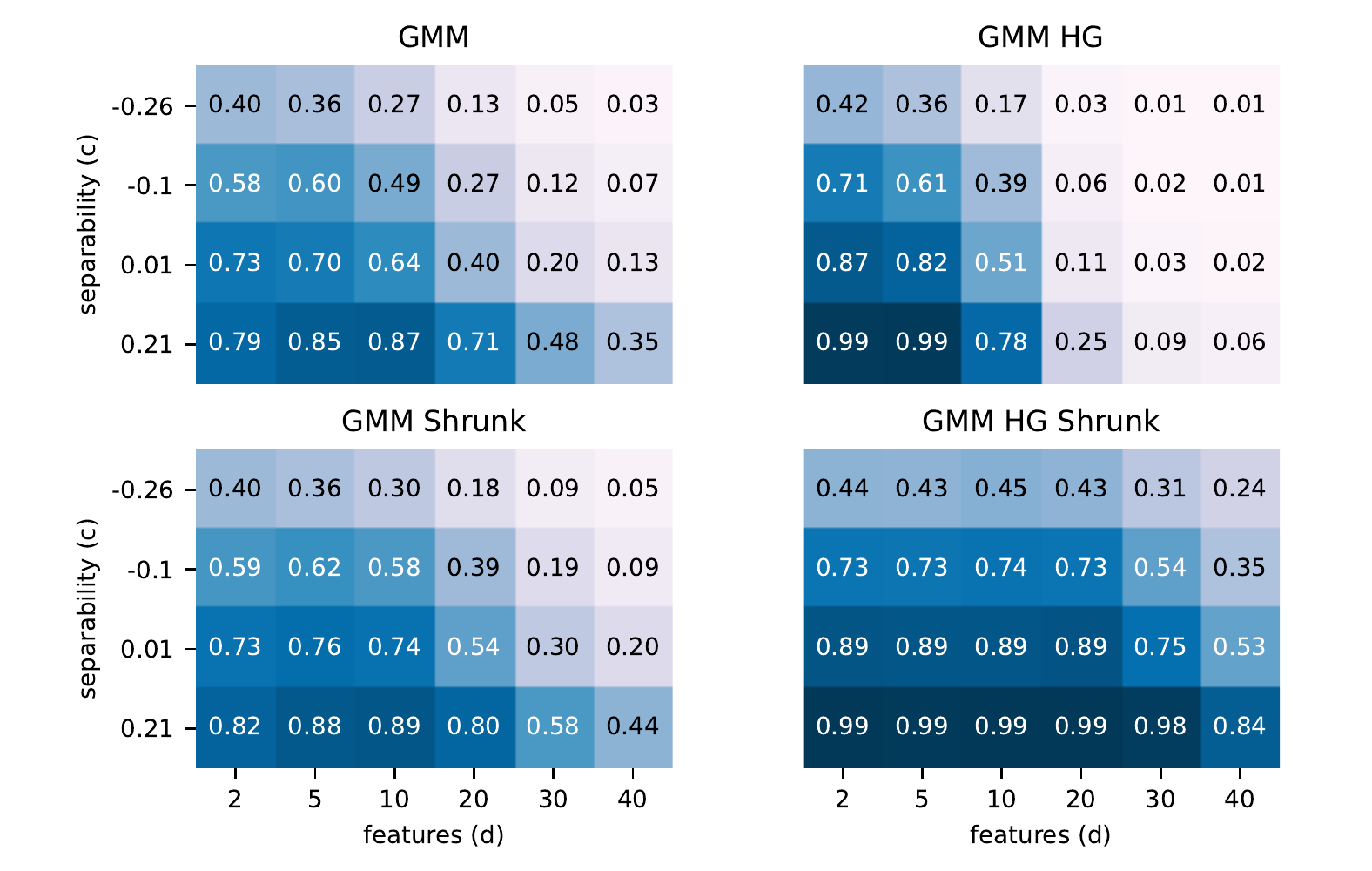}
\caption{Average ARI performance of GMM, GMM HG, GMM Shrunk, and GMM HG Shrunk over the $k = \{3, 10, 20\}$ datasets, for different separability indices and number of features}
\label{fig:heatmap8}
\end{figure}

\begin{figure}[htbp]
\centering
\includegraphics[width=0.85\textwidth]{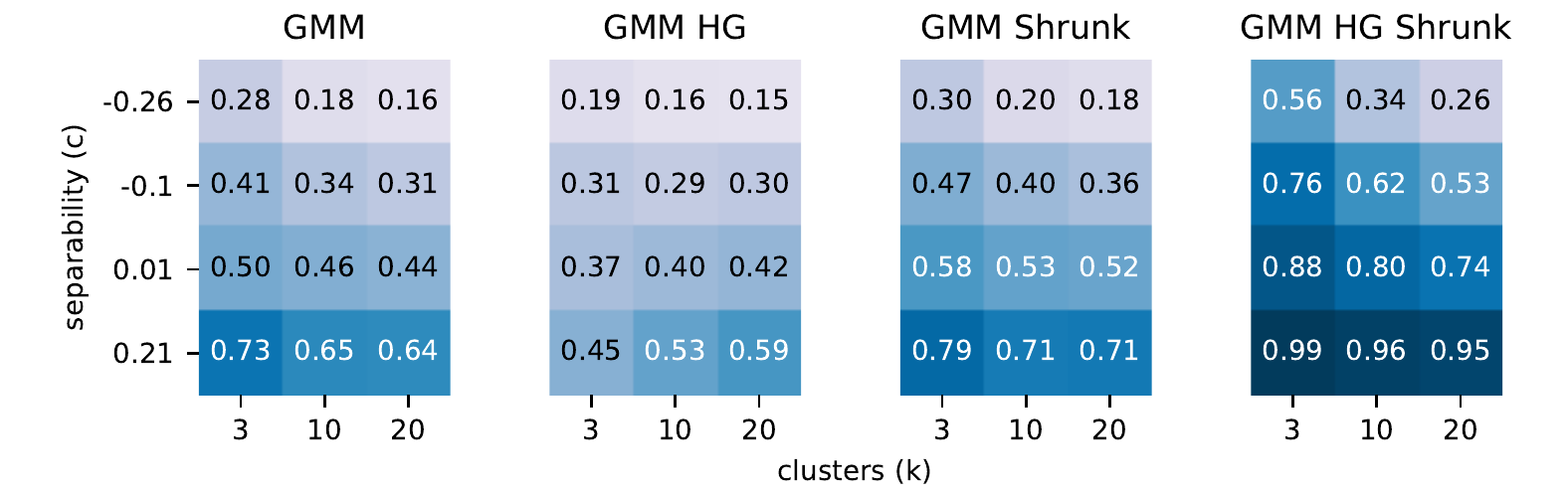}
\caption{Average ARI performance of GMM, GMM HG, GMM Shrunk, and GMM HG Shrunk over the $d = \{2, 5, 10, 20, 30, 40\}$ datasets, for different separability indices and number of clusters}
\label{fig:heatmap10}
\end{figure}

\figureref{fig:heatmap8}~and~\ref{fig:heatmap10} additionally report the results of GMM, GMM HG, GMM Shrunk, and GMM HG Shrunk for varying separability levels~$c$, number of clusters $k$, and number of features~$d$. Each heatmap gives the results of one of the four aforementioned search strategies. In both figures, the rows correspond to different levels of separability. In \figureref{fig:heatmap8}, each column corresponds to a different number of features, whereas in \figureref{fig:heatmap10} each column corresponds to a different number of clusters.

As seen in these results, the use of regularization improves the performance of the baseline GMM when the number of features increases. In contrast, the use of the HG search component for a more intensive search is, without regularization, only beneficial for problems of relatively-low dimension (e.g., it permits near-perfect cluster recovery with ARI of 0.99 compared to 0.85 when $c=0.21$ and $d=5$), and largely detrimental for higher-dimensional cases. 

Now, as seen in the results located in the bottom-right of \figureref{fig:heatmap8} and right of \figureref{fig:heatmap10}, using the Shrunk regularization and the HG search strategy jointly leads to a clear-cut improvement in all the considered situations, without any form of compromise. The regularization strategy effectively avoids overfitting in HG and permits fully harnessing its capabilities to achieve good clustering results in all regimes. Therefore, the results of this section indicate that combining GMM, regularization, and better search procedures is a promising approach for clustering.

\subsection{Comparisons with k-means and HG-means}
\label{sec:K-Means}

To broaden our analyses, we now include comparisons with two variants of the k-means algorithm. Indeed, k-means naturally lead to spherical clusters (due to the use of squared Euclidean distance in the objective of the underlying optimization problem) and can be viewed as a limit case of regularization. As the combination of k-means with HG search appeared to be a promising option in previous studies \citep{gribel2019hg}, it is meaningful to include additional comparisons of it with the newly proposed GMM HG algorithm with Shrunk regularization, also considering clustering performance and computational time.

\begin{figure}[htbp]
\centering
\includegraphics[width=0.9\textwidth]{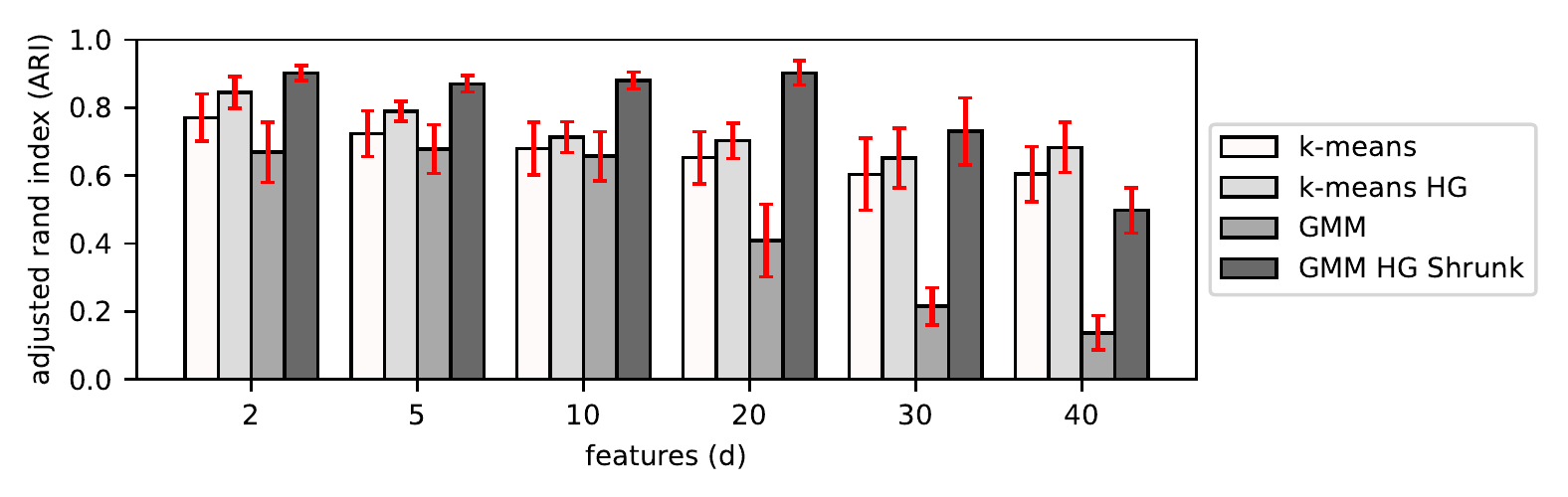}
\caption{Average and standard deviation of the ARI obtained by k-means, GMM and its \myblue{optimization} search variations over the $k = 10$ and $c = 0.01$ datasets}
\label{fig:optimization4}
\end{figure}

\figureref{fig:optimization4} therefore compare the clustering performance of a classical k-means, the Hybrid Genetic k-means as presented in \citet{gribel2019hg} (HG-means), a simple GMM, and the Hybrid Genetic GMM with regularization (GMM HG Shrunk). \figureref{fig:optimization4} present results for different values of $d$ with fixed values of the number of clusters and separability index ($k = 10$ and $c = 0.01$).

\begin{figure}[htbp]
\centering
\subfigure[Average of ARI measure]{
\includegraphics[width=0.85\textwidth]{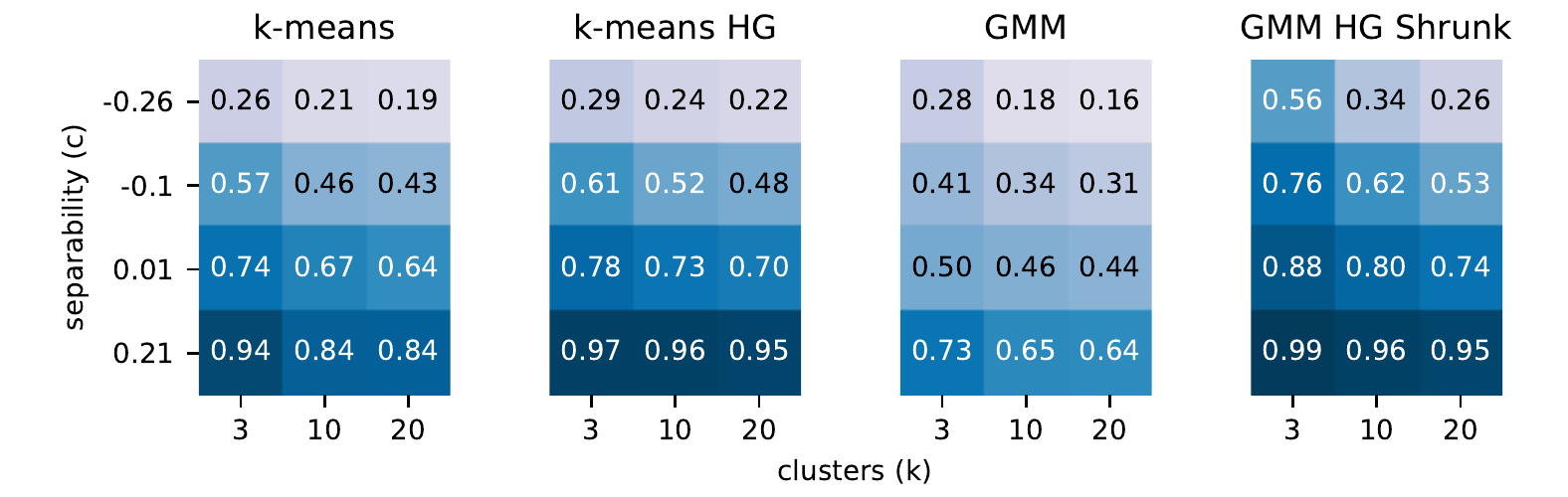}
\label{fig:heatmap5}
}
\vfill
\subfigure[Average of NMI measure]{
\includegraphics[width=0.85\textwidth]{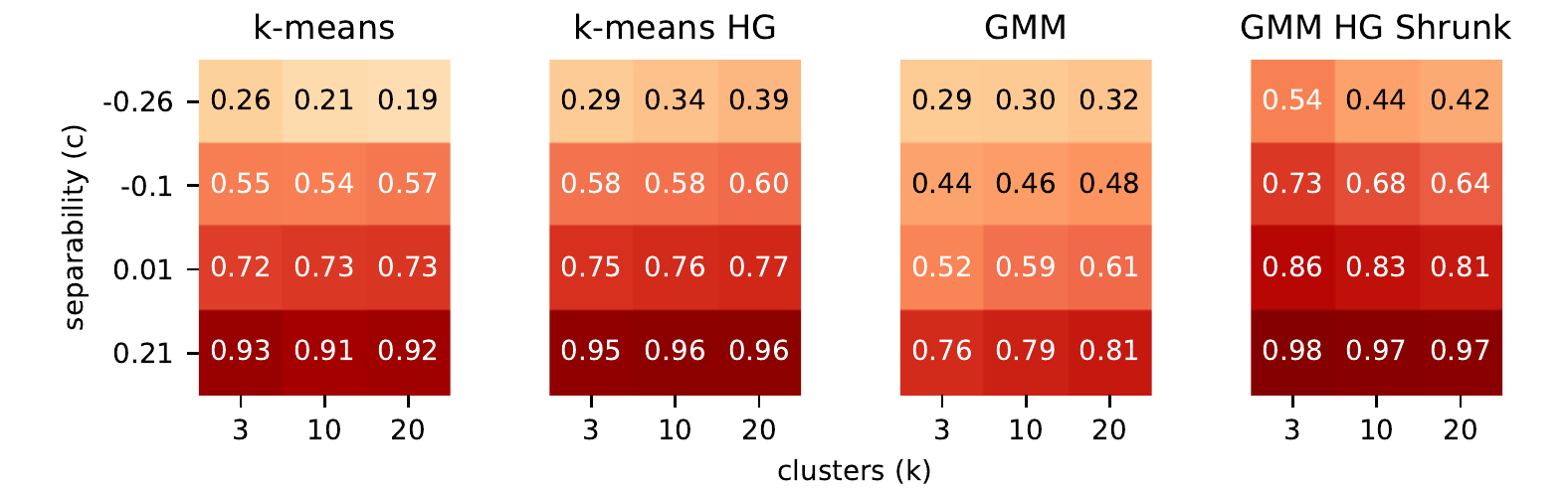}
\label{fig:heatmap12}
} 
\vfill
\subfigure[Average of CI measure]{
\includegraphics[width=0.85\textwidth]{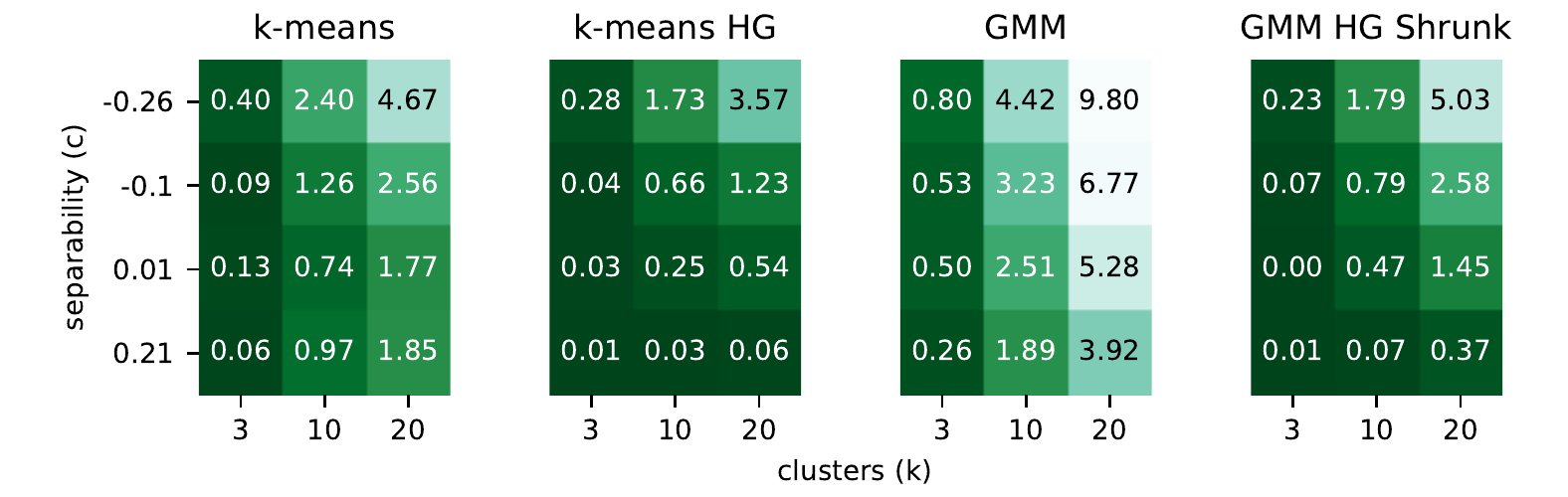}
\label{fig:heatmap11}
}
\caption{Performance of k-means, GMM, and its \myblue{optimization} search variations over the $d = \{2, 5, 10, 20, 30, 40\}$ datasets for different separability indices and number of clusters}
\label{fig:heatmap51211}
\end{figure}

\figureref{fig:heatmap51211} provides heatmaps that describe the evolution of the clustering performance when varying the separability index ($c$) and the number of clusters ($k$). \figureref{fig:heatmap5}, \figureref{fig:heatmap12}, \figureref{fig:heatmap11} present the results in terms of ARI, NMI, and CI, respectively. The format of these figures is the same as in \sectionref{sec:Regularization-Global}. Upon analysis, it is evident that the trends observed in the NMI and CI measures closely reflect those of the ARI. Specifically, when there is adequate \myblue{optimization}, the NMI approaches a value close to $1$, indicating high similarity between clusters, while the CI tends towards $0$, suggesting minimal clustering inconsistency. This consistency across the three metrics suggests that the clustering methods' performance remains relatively stable regardless of the specific evaluation measure used.

Both figures lead us to similar conclusions. Standard GMM systematically appears to be the worst method in terms of clustering quality, except when $c = -0.26$ and $k = 3$. Next, k-means ranks as the second-worst option, as the HG-based algorithms systematically outperform it. Finally, GMM HG Shrunk is generally superior to all other methods. This observation is confirmed by pairwise Wilcoxon tests between the pairs of methods presented in \tableref{tab:wilcoxon2}, considering the complete sets of results on all data sets.

\begin{table}[htbp]
\centering
\scalebox{0.9}
{
\begin{tabular}{l@{\hspace{1cm}}l}
\toprule
\textbf{Algorithm} & \textbf{p-value} \\
\midrule
GMM HG Shrunk -- k-means & $4.34\times 10^{-156}$ \\
GMM HG Shrunk -- k-means HG & $2.79\times 10^{-81}$ \\
GMM HG Shrunk -- GMM & $3.39\times 10^{-219}$ \\
\bottomrule
\end{tabular}
}
\caption{Pairwise Wilcoxon tests comparing the ARI of GMM HG Shrunk and the other algorithms}
\label{tab:wilcoxon2}
\end{table}

Interestingly, GMM HG Shrunk also achieved outstanding results compared to HG-means in instances with a low separability index.

Finally, \tableref{cputime1}~and~\ref{cputime2} compare the different algorithms' computational effort for subsets of the instances with a different number of clusters or features.

\begin{table}[htbp]
\centering
\scalebox{0.9}
{
\begin{tabular}{@{}c|rrrrrrrrrr@{}}
\toprule
\multirow{3}{*}{\#} & \multicolumn{1}{c}{k-means} & \multicolumn{1}{c}{k-means} & \multicolumn{1}{c}{GMM} & \multicolumn{1}{c}{GMM} & \multicolumn{1}{c}{GMM} & \multicolumn{1}{c}{GMM} & \multicolumn{1}{c}{GMM}   & \multicolumn{1}{c}{GMM}    & \multicolumn{1}{c}{GMM} & \multicolumn{1}{c}{GMM} \\
    & & \multicolumn{1}{c}{HG} & & \multicolumn{1}{c}{MS} & \multicolumn{1}{c}{RS}  & \multicolumn{1}{c}{HG} & & \multicolumn{1}{c}{MS} & \multicolumn{1}{c}{RS}     & \multicolumn{1}{c}{HG} \\
    & & & & & & & \multicolumn{1}{c}{Shrunk} & \multicolumn{1}{c}{Shrunk} & \multicolumn{1}{c}{Shrunk} & \multicolumn{1}{c}{Shrunk} \\ \midrule
3 & $0.01$ & $0.71$ & $0.01$ & $1.45$ & $1.02$ & $1.20$ & $0.01$ & $1.19$ & $0.72$ & $1.05$ \\
10 & $0.03$ & $3.88$ & $0.11$ & $19.65$ & $6.61$ & $15.20$ & $0.10$ & $17.50$ & $5.61$ & $12.52$ \\
20 & $0.07$ & $10.26$ & $0.42$ & $80.96$ & $21.22$ & $63.89$ & $0.41$ & $78.52$ & $19.19$ & $54.88$ \\
\midrule
Avg. & $0.04$ & $4.95$ & $0.18$ & $34.02$ & $9.62$ & $26.76$ & $0.17$ & $32.40$ & $8.51$ & $22.82$ \\ \bottomrule
\end{tabular}
}
\caption{Average CPU time in seconds for a different target number of clusters}
\label{cputime1}
\end{table}

\begin{table}[htbp]
\centering
\scalebox{0.9}
{
\begin{tabular}{@{}c|rrrrrrrrrr@{}}
\toprule
\multirow{3}{*}{\#} & \multicolumn{1}{c}{k-means} & \multicolumn{1}{c}{k-means} & \multicolumn{1}{c}{GMM} & \multicolumn{1}{c}{GMM} & \multicolumn{1}{c}{GMM} & \multicolumn{1}{c}{GMM} & \multicolumn{1}{c}{GMM}   & \multicolumn{1}{c}{GMM}    & \multicolumn{1}{c}{GMM} & \multicolumn{1}{c}{GMM} \\
    & & \multicolumn{1}{c}{HG} & & \multicolumn{1}{c}{MS} & \multicolumn{1}{c}{RS}  & \multicolumn{1}{c}{HG} & & \multicolumn{1}{c}{MS} & \multicolumn{1}{c}{RS}     & \multicolumn{1}{c}{HG} \\
    & & & & & & & \multicolumn{1}{c}{Shrunk} & \multicolumn{1}{c}{Shrunk} & \multicolumn{1}{c}{Shrunk} & \multicolumn{1}{c}{Shrunk} \\ \midrule
2 & $0.02$ & $3.59$ & $0.02$ & $3.93$ & $1.42$ & $3.25$ & $0.02$ & $3.44$ & $1.34$ & $2.97$ \\
5 & $0.02$ & $4.10$ & $0.05$ & $8.36$ & $3.17$ & $6.35$ & $0.04$ & $7.44$ & $2.72$ & $5.44$ \\
10 & $0.03$ & $4.40$ & $0.09$ & $16.50$ & $11.47$ & $12.73$ & $0.08$ & $14.70$ & $4.66$ & $9.77$ \\
20 & $0.03$ & $5.06$ & $0.22$ & $38.01$ & $12.72$ & $28.75$ & $0.17$ & $34.63$ & $8.65$ & $23.12$ \\
30 & $0.04$ & $6.03$ & $0.31$ & $62.70$ & $11.36$ & $49.42$ & $0.31$ & $60.91$ & $14.86$ & $43.03$ \\
40 & $0.04$ & $6.49$ & $0.37$ & $74.59$ & $17.54$ & $60.04$ & $0.38$ & $73.25$ & $18.79$ & $52.54$ \\
\midrule
Avg. & $0.03$ & $4.95$ & $0.18$ & $34.01$ & $9.61$ & $26.76$ & $0.17$ & $32.40$ & $8.50$ & $22.81$ \\ \bottomrule
\end{tabular}
}
\caption{\myblue{Average CPU time in seconds for a different target number of features}}
\label{cputime2}
\end{table}

As seen in these tables, k-means remains the fastest algorithm overall. Indeed, the steps of the method are very similar to GMM, but no covariance matrix estimation takes place, significantly decreasing the needed computational effort. 
More sophisticated search approaches such as MS, RS, or HG require several iterations of k-means or GMM, increasing their computational effort proportionally. Therefore, these methods are recommended in situations where computational effort is affordable.

\subsection{Performance on UCI Datasets}
\label{sec:UCI}

While synthetic data sets give fine-grained control over key factors ($n$, $c$, $d$, and~$k$), they do not provide the full picture of practical situations. In this section, we extend our experiments to datasets from the UCI repository \citep{Dua:2019} to observe whether the previous observations still apply. We gathered classical and recent datasets widely used in clustering benchmark comparisons. We focus on datasets containing a ground truth for properly assessing the clustering results. \tableref{uci_dimensions} lists the datasets labels and their dimensions ($n$, $d$, and~$k$). 

\begin{table}[htbp]
\centering
\scalebox{0.9}
{
\begin{tabular}{@{}c|l|rrr@{}}
\toprule
\multicolumn{1}{c|}{\#} & \multicolumn{1}{c|}{Dataset} & \multicolumn{1}{c}{\textit{n}} & \multicolumn{1}{c}{\textit{k}} & \multicolumn{1}{c}{\textit{d}} \\ \midrule
\texttt{A} & Ecoli \citep{misc_ecoli_39} & 336 & 8 & 7 \\
\texttt{B} & Facebook Live Sellers \citep{dehouche2020dataset} & 7050 & 4 & 9 \\
\texttt{C} & Fashion MNIST (Test) \citep{xiao2017online} & 10000 & 10 & 784 \\
\texttt{D} & Glass Identification \citep{misc_glass_identification_42} & 214 & 6 & 9 \\
\texttt{E} & HCV Data \citep{misc_hcv_data_571} & 589 & 5 & 12 \\
\texttt{F} & Human Activity Recognition \citep{misc_human_activity_recognition_using_smartphones_240} & 10299 & 6 & 561 \\
\texttt{G} & Image Segmentation \citep{misc_image_segmentation_50} & 2310 & 7 & 18 \\
\texttt{H} & Ionosphere \citep{misc_ionosphere_52} & 351 & 2 & 33 \\
\texttt{I} & Iris \citep{misc_iris_53} & 150 & 3 & 4 \\
\texttt{J} & Letter Recognition \citep{misc_letter_recognition_59} & 20000 & 26 & 16 \\
\texttt{K} & MAGIC Gamma Telescope \citep{misc_magic_gamma_telescope_159} & 19020 & 2 & 10 \\
\texttt{L} & Mice Protein Expression \citep{misc_mice_protein_expression_342} & 1047 & 8 & 74 \\
\texttt{M} & Optical Recognition of Handwritten Digits \citep{misc_optical_recognition_of_handwritten_digits_80} & 5620 & 10 & 64 \\
\texttt{N} & Pen-Based Recognition of Handwritten Digits \citep{misc_pen-based_recognition_of_handwritten_digits_81} & 10992 & 10 & 16 \\
\texttt{O} & Scadi \citep{misc_scadi_446} & 70 & 7 & 142 \\
\texttt{P} & Seeds \citep{misc_seeds_236} & 210 & 3 & 7 \\
\texttt{Q} & Soybean (Small) \citep{misc_soybean_(small)_91} & 47 & 4 & 21 \\
\texttt{R} & SPECT Heart \citep{misc_spect_heart_95} & 267 & 2 & 22 \\
\texttt{S} & Statlog (Heart) \citep{misc_statlog_(heart)_145} & 270 & 2 & 13 \\
\texttt{T} & Statlog (Shuttle) \citep{misc_statlog_(shuttle)_148} & 58000 & 7 & 9 \\
\texttt{U} & Waveform Database Generator (Version 1) \citep{misc_waveform_database_generator_(version_1)_107} & 5000 & 3 & 21 \\
\texttt{V} & Wholesale Customers \citep{misc_wholesale_customers_292} & 440 & 6 & 6 \\
\texttt{W} & Wine \citep{misc_wine_109} & 178 & 3 & 13 \\
\texttt{X} & Yeast \citep{misc_yeast_110} & 1484 & 10 & 8 \\
\texttt{Y} & Zoo \citep{misc_zoo_111} & 101 & 7 & 16 \\
\bottomrule
\end{tabular}
}
\caption{UCI datasets labels and dimensions}
\label{uci_dimensions}
\end{table}

The subsequent tables are compared to obtain a complete picture of the following methods: standard k-means, k-means HG (HG-means), standard GMM, GMM MS, GMM RS, GMM HG, GMM Shrunk, GMM MS Shrunk, GMM RS Shrunk, and GMM HG Shrunk. Since all algorithms depend on pseudo-random choices, ten runs of each algorithm have been done with a different random seed on each instance to improve statistical significance. Therefore, each table reports an average measurement over these ten runs. \myblue{For these experiments, we set the local search algorithms with a tolerance of 0.1 and capped their maximum iterations at 100.} We also highlight in each table the best performance in bold and use shades of grey to depict better results (darker is better). \tableref{uci_ari} reports each dataset's clustering accuracy (average of ARI).

\begin{table}[htbp]
\centering
\scalebox{0.9}
{
\begin{tabular}{@{}c|cccccccccc@{}}
\toprule
\multirow{3}{*}{\#} & k-means & k-means & GMM & GMM & GMM & GMM & GMM   & GMM    & GMM    & GMM    \\
                    &         & HG      &     & MS  & RS  & HG  &        & MS     & RS     & HG     \\
                    &         &         &     &     &     &     & Shrunk & Shrunk & Shrunk & Shrunk \\ \midrule

\texttt{A} & \cellcolor[HTML]{DDDDDD}0.40 & \cellcolor[HTML]{D7D7D7}0.46 & \cellcolor[HTML]{EEEEEE}0.22 & \cellcolor[HTML]{FFFFFF}0.04 & \cellcolor[HTML]{EFEFEF}0.21 & \cellcolor[HTML]{FFFFFF}0.04 & \cellcolor[HTML]{BFBFBF}\textbf{0.71} & \cellcolor[HTML]{BFBFBF}\textbf{0.71} & \cellcolor[HTML]{C0C0C0}0.70 & \cellcolor[HTML]{C1C1C1}0.69 \\
\texttt{B} & \cellcolor[HTML]{FAFAFA}0.10 & \cellcolor[HTML]{FFFFFF}0.09 & \cellcolor[HTML]{DFDFDF}0.16 & \cellcolor[HTML]{FAFAFA}0.10 & \cellcolor[HTML]{EDEDED}0.13 & \cellcolor[HTML]{FAFAFA}0.10 & \cellcolor[HTML]{BFBFBF}\textbf{0.23} & \cellcolor[HTML]{E8E8E8}0.14 & \cellcolor[HTML]{BFBFBF}\textbf{0.23} & \cellcolor[HTML]{E4E4E4}0.15 \\
\texttt{C} & \cellcolor[HTML]{C6C6C6}0.36 & \cellcolor[HTML]{CCCCCC}0.35 & \cellcolor[HTML]{F9F9F9}0.28 & \cellcolor[HTML]{E6E6E6}0.31 & \cellcolor[HTML]{FFFFFF}0.27 & \cellcolor[HTML]{FFFFFF}0.27 & \cellcolor[HTML]{D9D9D9}0.33 & \cellcolor[HTML]{BFBFBF}\textbf{0.37} & \cellcolor[HTML]{D9D9D9}0.33 & \cellcolor[HTML]{C6C6C6}0.36 \\
\texttt{D} & \cellcolor[HTML]{C4C4C4}0.25 & \cellcolor[HTML]{BFBFBF}\textbf{0.27} & \cellcolor[HTML]{CFCFCF}0.21 & \cellcolor[HTML]{FFFFFF}0.02 & \cellcolor[HTML]{CFCFCF}0.21 & \cellcolor[HTML]{F5F5F5}0.06 & \cellcolor[HTML]{BFBFBF}\textbf{0.27} & \cellcolor[HTML]{BFBFBF}\textbf{0.27} & \cellcolor[HTML]{BFBFBF}\textbf{0.27} & \cellcolor[HTML]{BFBFBF}\textbf{0.27} \\
\texttt{E} & \cellcolor[HTML]{D2D2D2}0.39 & \cellcolor[HTML]{BFBFBF}\textbf{0.47} & \cellcolor[HTML]{FFFFFF}0.20 & \cellcolor[HTML]{FDFDFD}0.21 & \cellcolor[HTML]{FFFFFF}0.20 & \cellcolor[HTML]{FDFDFD}0.21 & \cellcolor[HTML]{D7D7D7}0.37 & \cellcolor[HTML]{FFFFFF}0.20 & \cellcolor[HTML]{C6C6C6}0.44 & \cellcolor[HTML]{FAFAFA}0.22 \\
\texttt{F} & \cellcolor[HTML]{CACACA}0.49 & \cellcolor[HTML]{CFCFCF}0.47 & \cellcolor[HTML]{F4F4F4}0.33 & \cellcolor[HTML]{FCFCFC}0.30 & \cellcolor[HTML]{F7F7F7}0.32 & \cellcolor[HTML]{FFFFFF}0.29 & \cellcolor[HTML]{F2F2F2}0.34 & \cellcolor[HTML]{BFBFBF}\textbf{0.53} & \cellcolor[HTML]{DFDFDF}0.41 & \cellcolor[HTML]{C5C5C5}0.51 \\
\texttt{G} & \cellcolor[HTML]{DDDDDD}0.35 & \cellcolor[HTML]{DADADA}0.36 & \cellcolor[HTML]{FFFFFF}0.25 & \cellcolor[HTML]{F5F5F5}0.28 & \cellcolor[HTML]{F5F5F5}0.28 & \cellcolor[HTML]{FCFCFC}0.26 & \cellcolor[HTML]{DDDDDD}0.35 & \cellcolor[HTML]{BFBFBF}\textbf{0.44} & \cellcolor[HTML]{D0D0D0}0.39 & \cellcolor[HTML]{C6C6C6}0.42 \\
\texttt{H} & \cellcolor[HTML]{FFFFFF}0.18 & \cellcolor[HTML]{FFFFFF}0.18 & \cellcolor[HTML]{EEEEEE}0.33 & \cellcolor[HTML]{C6C6C6}0.69 & \cellcolor[HTML]{CDCDCD}0.63 & \cellcolor[HTML]{CFCFCF}0.61 & \cellcolor[HTML]{D4D4D4}0.56 & \cellcolor[HTML]{BFBFBF}\textbf{0.75} & \cellcolor[HTML]{C0C0C0}0.74 & \cellcolor[HTML]{C1C1C1}0.73 \\
\texttt{I} & \cellcolor[HTML]{FCFCFC}0.61 & \cellcolor[HTML]{E5E5E5}0.74 & \cellcolor[HTML]{FFFFFF}0.59 & \cellcolor[HTML]{D2D2D2}0.85 & \cellcolor[HTML]{E0E0E0}0.77 & \cellcolor[HTML]{CFCFCF}0.87 & \cellcolor[HTML]{FCFCFC}0.61 & \cellcolor[HTML]{D2D2D2}0.85 & \cellcolor[HTML]{C1C1C1}0.95 & \cellcolor[HTML]{BFBFBF}\textbf{0.96} \\
\texttt{J} & \cellcolor[HTML]{EAEAEA}0.14 & \cellcolor[HTML]{EEEEEE}0.13 & \cellcolor[HTML]{DDDDDD}0.17 & \cellcolor[HTML]{FBFBFB}0.10 & \cellcolor[HTML]{EAEAEA}0.14 & \cellcolor[HTML]{FFFFFF}0.09 & \cellcolor[HTML]{D4D4D4}0.19 & \cellcolor[HTML]{C8C8C8}0.22 & \cellcolor[HTML]{C8C8C8}0.22 & \cellcolor[HTML]{BFBFBF}\textbf{0.24} \\
\texttt{K} & \cellcolor[HTML]{FFFFFF}0.06 & \cellcolor[HTML]{FFFFFF}0.06 & \cellcolor[HTML]{EEEEEE}0.09 & \cellcolor[HTML]{FFFFFF}0.06 & \cellcolor[HTML]{FFFFFF}0.06 & \cellcolor[HTML]{FFFFFF}0.06 & \cellcolor[HTML]{BFBFBF}\textbf{0.17} & \cellcolor[HTML]{DCDCDC}0.12 & \cellcolor[HTML]{DCDCDC}0.12 & \cellcolor[HTML]{DCDCDC}0.12 \\
\texttt{L} & \cellcolor[HTML]{E1E1E1}0.52 & \cellcolor[HTML]{DADADA}0.57 & \cellcolor[HTML]{FFFFFF}0.33 & \cellcolor[HTML]{E9E9E9}0.47 & \cellcolor[HTML]{FFFFFF}0.33 & \cellcolor[HTML]{D8D8D8}0.58 & \cellcolor[HTML]{EFEFEF}0.43 & \cellcolor[HTML]{D8D8D8}0.58 & \cellcolor[HTML]{C2C2C2}0.72 & \cellcolor[HTML]{BFBFBF}\textbf{0.74} \\
\texttt{M} & \cellcolor[HTML]{D0D0D0}0.64 & \cellcolor[HTML]{CBCBCB}0.68 & \cellcolor[HTML]{EFEFEF}0.38 & \cellcolor[HTML]{FAFAFA}0.29 & \cellcolor[HTML]{FFFFFF}0.25 & \cellcolor[HTML]{FAFAFA}0.29 & \cellcolor[HTML]{CFCFCF}0.65 & \cellcolor[HTML]{BFBFBF}\textbf{0.78} & \cellcolor[HTML]{C3C3C3}0.75 & \cellcolor[HTML]{C0C0C0}0.77 \\
\texttt{N} & \cellcolor[HTML]{D6D6D6}0.54 & \cellcolor[HTML]{D6D6D6}0.54 & \cellcolor[HTML]{E8E8E8}0.40 & \cellcolor[HTML]{FFFFFF}0.23 & \cellcolor[HTML]{FBFBFB}0.26 & \cellcolor[HTML]{FEFEFE}0.24 & \cellcolor[HTML]{D6D6D6}0.54 & \cellcolor[HTML]{C9C9C9}0.64 & \cellcolor[HTML]{BFBFBF}\textbf{0.71} & \cellcolor[HTML]{C1C1C1}0.70 \\
\texttt{O} & \cellcolor[HTML]{BFBFBF}\textbf{0.31} & \cellcolor[HTML]{C8C8C8}0.30 & \cellcolor[HTML]{FFFFFF}0.24 & \cellcolor[HTML]{EDEDED}0.26 & \cellcolor[HTML]{FFFFFF}0.24 & \cellcolor[HTML]{E4E4E4}0.27 & \cellcolor[HTML]{F6F6F6}0.25 & \cellcolor[HTML]{D1D1D1}0.29 & \cellcolor[HTML]{BFBFBF}\textbf{0.31} & \cellcolor[HTML]{BFBFBF}\textbf{0.31} \\
\texttt{P} & \cellcolor[HTML]{C3C3C3}0.72 & \cellcolor[HTML]{C3C3C3}0.72 & \cellcolor[HTML]{E7E7E7}0.63 & \cellcolor[HTML]{FBFBFB}0.58 & \cellcolor[HTML]{FFFFFF}0.57 & \cellcolor[HTML]{F7F7F7}0.59 & \cellcolor[HTML]{E3E3E3}0.64 & \cellcolor[HTML]{C7C7C7}0.71 & \cellcolor[HTML]{BFBFBF}\textbf{0.73} & \cellcolor[HTML]{C3C3C3}0.72 \\
\texttt{Q} & \cellcolor[HTML]{EDEDED}0.62 & \cellcolor[HTML]{F5F5F5}0.55 & \cellcolor[HTML]{FDFDFD}0.49 & \cellcolor[HTML]{CACACA}0.90 & \cellcolor[HTML]{FFFFFF}0.47 & \cellcolor[HTML]{DDDDDD}0.75 & \cellcolor[HTML]{FAFAFA}0.51 & \cellcolor[HTML]{C7C7C7}0.93 & \cellcolor[HTML]{EEEEEE}0.61 & \cellcolor[HTML]{BFBFBF}\textbf{0.99} \\
\texttt{R} & \cellcolor[HTML]{F9F9F9}0.01 & \cellcolor[HTML]{FBFBFB}-0.00 & \cellcolor[HTML]{FFFFFF}-0.02 & \cellcolor[HTML]{F4F4F4}0.03 & \cellcolor[HTML]{DFDFDF}0.13 & \cellcolor[HTML]{E6E6E6}0.10 & \cellcolor[HTML]{FBFBFB}-0.00 & \cellcolor[HTML]{F2F2F2}0.04 & \cellcolor[HTML]{BFBFBF}\textbf{0.28} & \cellcolor[HTML]{C1C1C1}0.27 \\
\texttt{S} & \cellcolor[HTML]{E1E1E1}0.20 & \cellcolor[HTML]{BFBFBF}\textbf{0.35} & \cellcolor[HTML]{FAFAFA}0.09 & \cellcolor[HTML]{FFFFFF}0.07 & \cellcolor[HTML]{FFFFFF}0.07 & \cellcolor[HTML]{EFEFEF}0.14 & \cellcolor[HTML]{F4F4F4}0.12 & \cellcolor[HTML]{FAFAFA}0.09 & \cellcolor[HTML]{FDFDFD}0.08 & \cellcolor[HTML]{E1E1E1}0.20 \\
\texttt{T} & \cellcolor[HTML]{F2F2F2}0.30 & \cellcolor[HTML]{BFBFBF}\textbf{0.89} & \cellcolor[HTML]{F4F4F4}0.28 & \cellcolor[HTML]{FCFCFC}0.19 & \cellcolor[HTML]{FCFCFC}0.19 & \cellcolor[HTML]{FFFFFF}0.15 & \cellcolor[HTML]{E6E6E6}0.44 & \cellcolor[HTML]{E9E9E9}0.41 & \cellcolor[HTML]{E6E6E6}0.44 & \cellcolor[HTML]{E8E8E8}0.42 \\
\texttt{U} & \cellcolor[HTML]{FFFFFF}0.26 & \cellcolor[HTML]{FFFFFF}0.26 & \cellcolor[HTML]{F6F6F6}0.30 & \cellcolor[HTML]{BFBFBF}\textbf{0.53} & \cellcolor[HTML]{EEEEEE}0.33 & \cellcolor[HTML]{CBCBCB}0.48 & \cellcolor[HTML]{F6F6F6}0.30 & \cellcolor[HTML]{C2C2C2}0.52 & \cellcolor[HTML]{D2D2D2}0.45 & \cellcolor[HTML]{CDCDCD}0.47 \\
\texttt{V} & \cellcolor[HTML]{D4D4D4}0.14 & \cellcolor[HTML]{FFFFFF}0.12 & \cellcolor[HTML]{D4D4D4}0.14 & \cellcolor[HTML]{D4D4D4}0.14 & \cellcolor[HTML]{EAEAEA}0.13 & \cellcolor[HTML]{EAEAEA}0.13 & \cellcolor[HTML]{BFBFBF}\textbf{0.15} & \cellcolor[HTML]{EAEAEA}0.13 & \cellcolor[HTML]{BFBFBF}\textbf{0.15} & \cellcolor[HTML]{D4D4D4}0.14 \\
\texttt{W} & \cellcolor[HTML]{CACACA}0.37 & \cellcolor[HTML]{C8C8C8}0.38 & \cellcolor[HTML]{BFBFBF}\textbf{0.42} & \cellcolor[HTML]{D4D4D4}0.32 & \cellcolor[HTML]{FFFFFF}0.12 & \cellcolor[HTML]{EAEAEA}0.22 & \cellcolor[HTML]{D4D4D4}0.32 & \cellcolor[HTML]{D4D4D4}0.32 & \cellcolor[HTML]{CECECE}0.35 & \cellcolor[HTML]{D2D2D2}0.33 \\
\texttt{X} & \cellcolor[HTML]{D3D3D3}0.13 & \cellcolor[HTML]{CBCBCB}0.15 & \cellcolor[HTML]{FFFFFF}0.02 & \cellcolor[HTML]{EBEBEB}0.07 & \cellcolor[HTML]{FBFBFB}0.03 & \cellcolor[HTML]{EBEBEB}0.07 & \cellcolor[HTML]{CFCFCF}0.14 & \cellcolor[HTML]{C3C3C3}0.17 & \cellcolor[HTML]{CBCBCB}0.15 & \cellcolor[HTML]{BFBFBF}\textbf{0.18} \\
\texttt{Y} & \cellcolor[HTML]{C4C4C4}0.69 & \cellcolor[HTML]{C4C4C4}0.69 & \cellcolor[HTML]{D1D1D1}0.64 & \cellcolor[HTML]{BFBFBF}\textbf{0.71} & \cellcolor[HTML]{FFFFFF}0.46 & \cellcolor[HTML]{C4C4C4}0.69 & \cellcolor[HTML]{CFCFCF}0.65 & \cellcolor[HTML]{BFBFBF}\textbf{0.71} & \cellcolor[HTML]{FCFCFC}0.47 & \cellcolor[HTML]{E0E0E0}0.58 \\
\midrule
\multicolumn{1}{c|}{Avg.} & \cellcolor[HTML]{E4E4E4}0.35 & \cellcolor[HTML]{D7D7D7}0.39 & \cellcolor[HTML]{F8F8F8}0.29 & \cellcolor[HTML]{F2F2F2}0.31 & \cellcolor[HTML]{FFFFFF}0.27 & \cellcolor[HTML]{F5F5F5}0.30 & \cellcolor[HTML]{DDDDDD}0.37 & \cellcolor[HTML]{C6C6C6}0.44 & \cellcolor[HTML]{C6C6C6}0.44 & \cellcolor[HTML]{BFBFBF}\textbf{0.46} \\
\bottomrule
\end{tabular}
}
\caption{Accuracy (ARI) comparison on the UCI datasets}
\label{uci_ari}
\end{table}

\tableref{uci_metrics} presents a comprehensive comparison using the ARI, NMI, and CI metrics. These metrics provide insights into the accuracy and information retention of the clusters formed by each method. ). Notably, the NMI and CI metrics corroborate the findings from the ARI results, emphasizing the consistency in the performance evaluation across different metrics. These results indicate that the proposed methods offer superior \myblue{optimization} search capabilities, leading to more accurate and cohesive clustering outcomes. The consistency across all three metrics underscores the robustness of the proposed methods and their potential in handling diverse datasets.

\begin{table}[htbp]
\centering
\scalebox{0.9}
{
\begin{tabular}{@{}c|cccccccccc@{}}
\toprule
\multirow{3}{*}{Average} & k-means & k-means & GMM & GMM & GMM & GMM & GMM   & GMM    & GMM    & GMM    \\
                    &         & HG      &     & MS  & RS  & HG  &        & MS     & RS     & HG     \\
                    &         &         &     &     &     &     & Shrunk & Shrunk & Shrunk & Shrunk \\ \midrule
\multicolumn{1}{c|}{ARI} & \cellcolor[HTML]{E4E4E4}0.35 & \cellcolor[HTML]{D7D7D7}0.39 & \cellcolor[HTML]{F8F8F8}0.29 & \cellcolor[HTML]{F2F2F2}0.31 & \cellcolor[HTML]{FFFFFF}0.27 & \cellcolor[HTML]{F5F5F5}0.30 & \cellcolor[HTML]{DDDDDD}0.37 & \cellcolor[HTML]{C6C6C6}0.44 & \cellcolor[HTML]{C6C6C6}0.44 & \cellcolor[HTML]{BFBFBF}\textbf{0.46} \\
\multicolumn{1}{c|}{NMI} & \cellcolor[HTML]{E1E1E1}0.44 & \cellcolor[HTML]{DDDDDD}0.45 & \cellcolor[HTML]{FBFBFB}0.38 & \cellcolor[HTML]{F2F2F2}0.40 & \cellcolor[HTML]{FFFFFF}0.37 & \cellcolor[HTML]{F6F6F6}0.39 & \cellcolor[HTML]{E1E1E1}0.44 & \cellcolor[HTML]{C4C4C4}0.51 & \cellcolor[HTML]{C8C8C8}0.50 & \cellcolor[HTML]{BFBFBF}\textbf{0.52} \\
\multicolumn{1}{c|}{CI} & \cellcolor[HTML]{DCDCDC}2.28 & \cellcolor[HTML]{C7C7C7}2.09 & \cellcolor[HTML]{FFFFFF}2.60 & \cellcolor[HTML]{E7E7E7}2.38 & \cellcolor[HTML]{FFFFFF}2.60 & \cellcolor[HTML]{F1F1F1}2.47 & \cellcolor[HTML]{F2F2F2}2.48 & \cellcolor[HTML]{C7C7C7}2.09 & \cellcolor[HTML]{E4E4E4}2.35 & \cellcolor[HTML]{BFBFBF}\textbf{2.02} \\
\bottomrule
\end{tabular}
}
\caption{Comparison of the ARI, NMI, and CI averages on the UCI datasets}
\label{uci_metrics}
\end{table}

\tableref{uci_cpu} reports the CPU time average of the ten runs for each algorithm on each dataset.

\begin{table}[htbp]
\centering
\scalebox{0.9}
{
\begin{tabular}{@{}c|rrrrrrrrrr@{}}
\toprule
\multirow{3}{*}{\#} & \multicolumn{1}{c}{k-means} & \multicolumn{1}{c}{k-means} & \multicolumn{1}{c}{GMM} & \multicolumn{1}{c}{GMM} & \multicolumn{1}{c}{GMM} & \multicolumn{1}{c}{GMM} & \multicolumn{1}{c}{GMM}   & \multicolumn{1}{c}{GMM}    & \multicolumn{1}{c}{GMM} & \multicolumn{1}{c}{GMM} \\
    & & \multicolumn{1}{c}{HG} & & \multicolumn{1}{c}{MS} & \multicolumn{1}{c}{RS}  & \multicolumn{1}{c}{HG} & & \multicolumn{1}{c}{MS} & \multicolumn{1}{c}{RS}     & \multicolumn{1}{c}{HG} \\
    & & & & & & & \multicolumn{1}{c}{Shrunk} & \multicolumn{1}{c}{Shrunk} & \multicolumn{1}{c}{Shrunk} & \multicolumn{1}{c}{Shrunk} \\ \midrule
\texttt{A}& $0.01$ & $0.39$ & $0.03$ & $3.10$ & $1.12$ & $6.39$ & $0.01$ & $0.78$ & $0.28$ & $0.54$ \\
\texttt{B}& $0.26$ & $41.19$ & $0.07$ & $10.21$ & $6.03$ & $7.87$ & $0.06$ & $9.51$ & $3.06$ & $6.62$ \\
\texttt{C}& $3.30$ & $733.20$ & $75.02$ & $14217.75$ & $4300.53$ & $9248.45$ & $69.44$ & $15752.81$ & $8761.73$ & $13723.37$ \\
\texttt{D}& $0.01$ & $0.38$ & $0.02$ & $4.09$ & $2.23$ & $10.67$ & $0.01$ & $0.30$ & $0.23$ & $0.31$ \\
\texttt{E}& $0.02$ & $2.84$ & $0.02$ & $0.99$ & $0.86$ & $1.26$ & $0.01$ & $0.61$ & $0.34$ & $0.50$ \\
\texttt{F}& $2.88$ & $603.91$ & $24.15$ & $4259.31$ & $1562.01$ & $3101.94$ & $17.86$ & $3538.06$ & $1335.69$ & $2860.72$ \\
\texttt{G}& $0.08$ & $11.56$ & $0.11$ & $19.93$ & $6.97$ & $12.82$ & $0.03$ & $4.63$ & $3.00$ & $3.88$ \\
\texttt{H}& $0.01$ & $0.54$ & $0.01$ & $0.84$ & $0.65$ & $0.91$ & $0.01$ & $0.67$ & $0.42$ & $0.57$ \\
\texttt{I}& $0.01$ & $0.16$ & $0.01$ & $0.10$ & $0.13$ & $0.09$ & $0.01$ & $0.09$ & $0.05$ & $0.09$ \\
\texttt{J}& $2.31$ & $390.99$ & $1.38$ & $279.70$ & $153.38$ & $479.43$ & $0.95$ & $194.66$ & $79.93$ & $154.44$ \\
\texttt{K}& $0.40$ & $71.39$ & $0.07$ & $9.13$ & $7.90$ & $9.57$ & $0.05$ & $7.86$ & $5.62$ & $7.81$ \\
\texttt{L}& $0.05$ & $7.47$ & $0.24$ & $44.44$ & $35.51$ & $36.96$ & $0.09$ & $16.20$ & $11.60$ & $17.69$ \\
\texttt{M}& $0.28$ & $79.46$ & $0.94$ & $201.90$ & $49.91$ & $123.49$ & $0.74$ & $150.84$ & $48.36$ & $93.99$ \\
\texttt{N}& $0.36$ & $79.11$ & $0.79$ & $166.78$ & $221.47$ & $279.06$ & $0.23$ & $45.57$ & $21.29$ & $37.16$ \\
\texttt{O}& $0.01$ & $0.26$ & $0.11$ & $19.37$ & $8.83$ & $18.61$ & $0.04$ & $7.10$ & $7.95$ & $9.38$ \\
\texttt{P}& $0.01$ & $0.33$ & $0.01$ & $0.17$ & $0.16$ & $0.17$ & $0.01$ & $0.12$ & $0.12$ & $0.15$ \\
\texttt{Q}& $0.01$ & $0.06$ & $0.01$ & $0.84$ & $0.52$ & $0.77$ & $0.01$ & $0.18$ & $0.21$ & $0.24$ \\
\texttt{R}& $0.01$ & $0.57$ & $0.01$ & $0.58$ & $0.32$ & $0.50$ & $0.01$ & $0.30$ & $0.19$ & $0.27$ \\
\texttt{S}& $0.01$ & $0.29$ & $0.01$ & $0.30$ & $0.23$ & $0.31$ & $0.01$ & $0.13$ & $0.09$ & $0.14$ \\
\texttt{T}& $2.50$ & $522.81$ & $0.76$ & $169.70$ & $75.20$ & $130.22$ & $0.40$ & $85.92$ & $34.16$ & $69.18$ \\
\texttt{U}& $0.06$ & $12.60$ & $0.03$ & $3.25$ & $2.85$ & $3.49$ & $0.02$ & $3.20$ & $3.05$ & $3.51$ \\
\texttt{V}& $0.01$ & $1.49$ & $0.01$ & $0.44$ & $0.34$ & $0.45$ & $0.01$ & $0.32$ & $0.17$ & $0.28$ \\
\texttt{W}& $0.01$ & $0.27$ & $0.01$ & $0.17$ & $0.25$ & $0.28$ & $0.01$ & $0.46$ & $0.56$ & $0.58$ \\
\texttt{X}& $0.02$ & $2.16$ & $0.03$ & $4.02$ & $2.51$ & $3.83$ & $0.02$ & $3.21$ & $0.90$ & $1.90$ \\
\texttt{Y}& $0.01$ & $0.12$ & $0.02$ & $1.17$ & $0.75$ & $1.17$ & $0.01$ & $0.33$ & $0.33$ & $0.43$ \\
\midrule
\multicolumn{1}{c|}{Avg.}& $0.50$ & $102.54$ & $4.15$ & $776.73$ & $257.63$ & $539.15$ & $3.60$ & $792.95$ & $412.77$ & $679.75$ \\ \bottomrule
\end{tabular}
}
\caption{CPU time (s) comparison on the UCI datasets}
\label{uci_cpu}
\end{table}

\begin{table}[htbp]
\centering
\scalebox{0.9}
{
\begin{tabular}{l@{\hspace{1cm}}l}
\toprule
\textbf{Algorithm} & \textbf{p-value} \\
\midrule
GMM HG Shrunk -- k-means & $1.19\times 10^{-19}$ \\
GMM HG Shrunk -- k-means HG & $1.64\times 10^{-10}$ \\
GMM HG Shrunk -- GMM & $1.20\times 10^{-28}$ \\
GMM HG Shrunk -- GMM MS & $1.87\times 10^{-28}$ \\
GMM HG Shrunk -- GMM RS & $5.25\times 10^{-37}$ \\
GMM HG Shrunk -- GMM HG & $2.17\times 10^{-31}$ \\
GMM HG Shrunk -- GMM Shrunk & $6.69\times 10^{-15}$ \\
GMM HG Shrunk -- GMM MS Shrunk & $1.86\times 10^{-06}$ \\
GMM HG Shrunk -- GMM RS Shrunk & $2.46\times 10^{-02}$ \\
\bottomrule
\end{tabular}
}
\caption{Pairwise Wilcoxon tests comparing the ARI of GMM HG Shrunk and the other algorithms}
\label{tab:wilcoxon3}
\end{table}

The results of this experiment confirm the observations from the previous sections. We see that the combination of GMM, regularization, and \myblue{optimization} consistently achieves the best performance. Overall, the regularized GMM with \myblue{an optimized} search strategy (i.e., GMM HG Shrunk, closely followed by GMM RS Shrunk) appears to be the best method overall: it achieves the best results or closely-tied results on most datasets. As seen in \tableref{tab:wilcoxon3}, the superiority of GMM HG Shrunk in terms of accuracy is confirmed at a 5\% confidence level by pairwise Wilcoxon tests (considering the individual results of all runs on all data sets) comparing the ARI of the GMM HG Shrunk method with that of the other algorithms.

\myblue{In terms of computational effort, most superior search algorithms require multiple search trajectories and, consequently, additional computational time. } Nevertheless, CPU time does not exceed a few minutes except on the \myblue{\textit{Fashion MNIST (Test)} (\texttt{C}), \textit{Human Activity Recognition} (\texttt{F}), and \textit{Letter Recognition} (\texttt{J}) datasets}. In those cases, if the computational effort becomes a bottleneck, parallel implementations of the proposed methods can be developed, e.g., by performing multiple solution generation by crossover and EM local search simultaneously.

Furthermore, it is essential to highlight that the methods discussed in this paper are not limited to the datasets mentioned but can also be effectively applied to image-related datasets. In fact, we have already tested the method on four specific image datasets: \myblue{\textit{Fashion MNIST (Test)} (\texttt{C}), \textit{Optical Recognition of Handwritten Digits} (\texttt{M}), \textit{Image Segmentation} (\texttt{G}), and \textit{Pen-Based Recognition of Handwritten Digits} (\texttt{N}).} While the \myblue{\textit{Letter Recognition} (\texttt{J})} dataset can be associated with image processing, it is important to note that it is not strictly an image dataset. The successful application of our method to these datasets underscores its versatility and potential for broader applications in the domain of image processing and related fields.

\section{Conclusion}
\label{section:conclusion}

Due to their conceptual simplicity, k-means algorithm variants have been extensively used for unsupervised clustering. However, these algorithms are limited in their capacity by the fact that they essentially fit spherical Gaussian distributions to data that significantly deviates from these distributional assumptions. Against this background, GMM approaches give more modeling flexibility, but their increased number of parameters leads to more challenging underlying optimization problems and possible overfit. 
    
In this work, we have revisited this status quo by examining the relations between regularization techniques and more sophisticated search methods within the GMM model, therefore circumventing both aforementioned weaknesses. We introduced an efficient population-based GMM (GMM HG) combined with simple regularization strategies. Through extensive numerical experiments on synthetic and real data, we demonstrate that this combination of \myblue{optimization} and proper regularization permits achieving a totally new level of performance in terms of clustering accuracy (ARI). Strikingly, the use of more advanced sophisticated optimization alone was not sufficient to achieve these results; and even proved detrimental to some high-dimensional datasets. In a similar fashion, the use of regularization alone did not have a significant impact on the performance of the classic GMM. It is truly the combination of both strategies that permitted a major performance breakthrough. 

This study opens many promising research avenues. First of all, we suggest pursuing the analysis of different regularization strategies in this context, e.g., to better learn or adapt the $\delta_{\text{Shrunk}}$ parameter. We also suggest further refining the solution methods to achieve faster and more accurate results. In particular, if computational time becomes a bottleneck, parallel computing may be used to generate multiple solutions simultaneously and speed up the solution process of HG or RS algorithmic variants. Finally, additional adaptations and specializations could be envisaged for different domains or data types, e.g., for computer vision or time-series analysis. To facilitate further studies, we provide all the data and analyses from our study in an open-source repository at \myblue{\url{http://www.github.com/raphasampaio/RegularizationAndOptimizationInModelBasedClustering.jl}}, as well the Julia packages \textsc{UnsupervisedClustering.jl} and \textsc{RegularizedCovarianceMatrices.jl}.

\section*{Acknowledgements}
The authors gratefully acknowledge PSR for providing all the cloud infrastructure and recognize the contributions of the team on the development that led to this work.

This research was partially supported by the Coordena\c c\~ao de Aperfei\c coamento de Pessoal de N\'ivel Superior - Brasil (CAPES) - Finance Code 001, by CNPq under grant 308528/2018-2, and by FAPERJ under grant E-26/202.790/2019 in Brazil. This financial support is gratefully acknowledged.

\bibliographystyle{ormsv080noURLDOI.bst}
\bibliography{bibliography.bib}

\end{document}